\documentclass[letterpaper, 10pt, conference]{ieeeconf}
\IEEEoverridecommandlockouts
\usepackage{cite}
\usepackage{amsmath,amssymb}
\usepackage{graphicx}
\usepackage{booktabs}
\usepackage{tikz}
\usetikzlibrary{positioning,arrows.meta,fit,calc}
\usepackage{pgfplots}
\pgfplotsset{compat=1.15}
\usepgfplotslibrary{groupplots}
\usetikzlibrary{backgrounds}

\definecolor{cbBlue}{RGB}{0,114,178}     
\definecolor{cbGreen}{RGB}{0,158,115}    
\definecolor{cbVerm}{RGB}{213,94,0}      
\definecolor{cbOrange}{RGB}{230,159,0}   
\definecolor{cbGray}{RGB}{115,115,115}   
\definecolor{fVis}{HTML}{E3F2EC}   \definecolor{sVis}{HTML}{2E8B6F}
\definecolor{fFor}{HTML}{FDEBD7}   \definecolor{sFor}{HTML}{E07B39}
\definecolor{fSta}{HTML}{EEF0F3}   \definecolor{sSta}{HTML}{7A8390}
\definecolor{fStu}{HTML}{E6EEF9}   \definecolor{sStu}{HTML}{3B6FB6}
\definecolor{fPri}{HTML}{F1E8F7}   \definecolor{sPri}{HTML}{8A5FB4}
\definecolor{fPan}{HTML}{FAFAFB}   \definecolor{sPan}{HTML}{D8DCE3}
\definecolor{tTitle}{HTML}{3C4048}
\tikzset{
  ffont/.style={font=\sffamily\scriptsize},
  card/.style={draw=#1!85!black, fill=#1!9, rounded corners=2.5pt,
               line width=0.7pt, inner sep=4pt, align=center,
               font=\sffamily\scriptsize},
  chip/.style={draw=#1!85!black, fill=#1!12, rounded corners=1.8pt,
               line width=0.55pt, inner sep=2.6pt, align=center,
               font=\sffamily\tiny},
  flow/.style={-{Stealth[length=1.9mm]}, line width=0.85pt, draw=#1},
  wire/.style={line width=0.55pt, draw=#1},
  lbl/.style={font=\sffamily\tiny, text=black!62, align=center},
  ptag/.style={font=\sffamily\bfseries\small},
}

\usepackage{multirow}
\usepackage{xcolor}
\usepackage{url}
\usepackage[hidelinks]{hyperref}

\newcommand{\ANONYMOUS}{0}   

\title{\LARGE\bf Seeing Through the Displaced Frame:\\
Privileged Noise Distillation for Vision-Force Precision Assembly}

\author{%
\if\ANONYMOUS1
  Anonymous Submission
\else
  Ching-Hsiang Chang$^{1,*}$ \quad Tzu-Yu Chuang$^{2}$ \quad Yi-Hsiu Lee$^{1}$ \quad
  Yi-Ting Chen$^{2}$ \quad Yuan-Fu Yang$^{2}$ \quad Min Sun$^{1}$%
  \thanks{$^{*}$Corresponding author. {\tt\small jimmy.chang@gapp.nthu.edu.tw}}%
  \thanks{$^{1}$National Tsing Hua University, Hsinchu, Taiwan.}%
  \thanks{$^{2}$National Yang Ming Chiao Tung University, Hsinchu, Taiwan.}%
\fi
}

\begin{document}
\maketitle
\thispagestyle{empty}
\pagestyle{empty}

\begin{abstract}
Pose error in precision assembly can corrupt not only what a robot observes but also the coordinate frame in which it acts. On the FORGE benchmark, the official state-based policy succeeds in $97\%$ to $99\%$ of episodes with the true pose but only $32\%$ to $60\%$ at the benchmark's $\sigma=5$\,mm pose-noise setting. The same estimated pose enters the observation and anchors the \emph{action frame}, making the offset unidentifiable from proprioceptive state alone before contact. We supply this missing information during training in two ways. A privileged teacher observes the offset in simulation, while clean demonstrations can instead be relabelled into the displaced frame in closed form. The deployed student is trained with behaviour cloning followed by one DAgger round and receives only noisy state, a raw wrench window, and two RGB cameras at test time. On the unmodified FORGE tasks, the teacher-route student maintains $92\%$ to $99\%$ success across $\sigma=0$ to $5$\,mm, while six non-privileged baselines fall to $2\%$ to $80\%$. A matched behaviour-cloning experiment isolates the source of this robustness. With the same student architecture, data budget, and training procedure, demonstrations generated without offset access yield only $31.5\%$ success at $\sigma=5$\,mm, whereas privileged and relabelled demonstrations reach $88.7\%$ and $95.8\%$. Deployed zero-shot on a Franka, the student reaches $83.3\%$ pooled success at $\sigma=5$\,mm against $34.4\%$ for the strongest state-based policy. Deployable sensing alone is insufficient. Robustness requires supervision that encodes compensation for the latent frame offset.
\end{abstract}

\section{Introduction}

\begin{figure}[t]
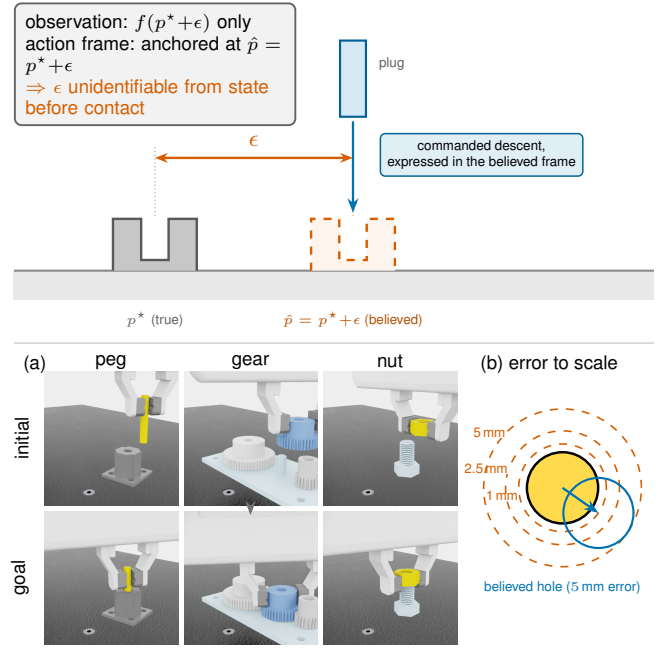

\centering
\resizebox{\columnwidth}{!}{%
\begin{tikzpicture}[ffont]
\fill[black!9] (0.15,0.52) rectangle (8.55,0.90);
\draw[black!45, line width=0.8pt] (0.15,0.90) -- (8.55,0.90);
\draw[black!65, line width=0.9pt, fill=black!22]
  (1.45,0.90) -- (1.45,1.58) -- (1.82,1.58) -- (1.82,1.04)
  -- (2.18,1.04) -- (2.18,1.58) -- (2.55,1.58) -- (2.55,0.90);
\draw[cbVerm, dashed, line width=0.9pt, fill=cbVerm!7]
  (4.05,0.90) -- (4.05,1.58) -- (4.42,1.58) -- (4.42,1.04)
  -- (4.78,1.04) -- (4.78,1.58) -- (5.15,1.58) -- (5.15,0.90);
\draw[black!40, densely dotted] (2.0,1.62) -- (2.0,2.52);
\draw[cbVerm!70, densely dotted] (4.6,1.62) -- (4.6,2.52);
\draw[{Stealth[length=1.7mm]}-{Stealth[length=1.7mm]}, cbVerm, line width=0.9pt]
  (2.0,2.38) -- node[above=0.5pt, font=\small\itshape, text=cbVerm] {$\epsilon$} (4.6,2.38);
\draw[cbBlue!85!black, line width=0.8pt, fill=cbBlue!16] (4.43,2.92) rectangle (4.77,3.92);
\node[lbl, anchor=west] at (4.82,3.62) {plug};
\draw[flow=cbBlue] (4.60,2.86) -- (4.60,1.66);
\node[chip=cbBlue, anchor=west] at (4.95,2.42)
  {commanded descent,\\ expressed in the believed frame};
\node[card=cbGray, anchor=north west, align=left, text width=3.4cm] at (0.15,4.42)
  {observation: $f(p^\star{+}\epsilon)$ only\\
   action frame: anchored at $\hat p = p^\star{+}\epsilon$\\
   \textcolor{cbVerm}{$\Rightarrow$ $\epsilon$ unidentifiable from state before contact}};
\node[lbl, anchor=north] at (2.0,0.48) {$p^\star$ (true)};
\node[lbl, anchor=north, text=cbVerm!85!black] at (4.6,0.48) {$\hat p = p^\star{+}\epsilon$ (believed)};
\draw[black!25] (0.15,-0.05) -- (8.55,-0.05);
\node[anchor=north west, font=\sffamily\scriptsize] at (0.15,-0.08) {(a)};
\foreach \t/\c/\lab in {peg/0/peg, gear/1/gear, nut/2/nut} {
  \node[anchor=north, font=\sffamily\scriptsize] at ({0.55+1.75/2+\c*1.83},-0.08) {\lab};
  \node[anchor=north west, inner sep=0] at ({0.55+\c*1.83},-0.42) {\includegraphics[width=1.75cm]{render/task_tiles/\t_initial.png}};
  \node[anchor=north west, inner sep=0] at ({0.55+\c*1.83},-2.25) {\includegraphics[width=1.75cm]{render/task_tiles/\t_goal.png}};
}
\node[anchor=south, font=\sffamily\scriptsize, rotate=90] at (0.47,-1.295) {initial};
\node[anchor=south, font=\sffamily\scriptsize, rotate=90] at (0.47,-3.125) {goal};
\draw[flow=black!55] (3.255,-2.19) -- (3.255,-2.31);
\node[anchor=north west, font=\sffamily\scriptsize] at (6.15,-0.08) {(b) error to scale};
\begin{scope}[shift={(7.35,-1.95)}, scale=0.115]
  \draw[black, fill=black!20, line width=0.6pt] (0,0) circle (4.114);
  \draw[black, fill=yellow!70!orange!80, line width=0.5pt] (0,0) circle (4.0);
  \foreach \r/\ang/\lab in {5/190/1\,mm, 6.5/160/2.5\,mm, 9/135/5\,mm} {
    \draw[cbVerm, dashed, line width=0.6pt] (0,0) circle (\r);
    \node[font=\sffamily\tiny, text=cbVerm, anchor=east, fill=white, inner sep=0.3pt] at (\ang:\r) {\lab};
  }
  \draw[cbBlue, line width=0.8pt] (-35:5) circle (4.0);
  \draw[flow=cbBlue] (0,0) -- (-35:5);
  \node[font=\sffamily\tiny, text=cbBlue, anchor=north, align=center] at (0,-9.6)
    {believed hole ($5$\,mm error)};
\end{scope}
\end{tikzpicture}}
\caption{The displaced action frame. \textbf{Top:} $\epsilon$ contaminates both the observation and action frame, making it unidentifiable from proprioceptive state before contact. \textbf{Bottom:} (a) initial and goal states of the three tasks. (b) Pose error to scale. At $5$\,mm, the displaced frame lies outside the peg hole.}
\label{fig:overview}
\end{figure}

Precision assembly policies can approach saturation in simulation when the fixture pose is accurately known. On the three FORGE tasks \cite{noseworthy2025forge}, peg insertion with a radial clearance of $0.114$\,mm, gear meshing, and M16 nut threading, the official policy reaches $97\%$ to $99\%$ success. Yet at $\sigma=5$\,mm of fixture-pose noise, the same policy falls to $31.8\%$, $49.9\%$, and $60.4\%$.

This degradation is not simply a perception failure. Section~\ref{sec:trap} identifies a more specific mechanism. In FORGE, as in standard task-frame formulations \cite{bruyninckx1996task}, the same estimated pose that enters the observation also anchors the \emph{action frame}. The policy commands displacement relative to the frame defined by that estimate. A single draw of pose noise shifts the observation and command frame together, making the underlying offset unidentifiable from proprioceptive state alone before contact. This coupling turns pose-estimation error into a control-frame offset that is latent to proprioception before contact.

We expose this offset during training and use it to supervise a policy that must compensate from deployable sensory cues at test time (Fig.~\ref{fig:overview}). We instantiate this supervision in two ways. A privileged teacher receives clean privileged state together with the offset and generates compensating demonstrations. When the displacement is a known translation, clean demonstrations from the \emph{official} policy can instead be relabelled into the displaced frame in closed form, removing the need for a privileged teacher. The deployed student uses only noisy state, a raw six-axis wrench history, and two RGB cameras, and is trained with behaviour cloning (BC) followed by one DAgger round. The target may be visible, yet visibility alone does not determine how the policy should compensate for the displaced action frame. A supervised frame corrector provides an additional control by estimating the offset from vision and correcting the frame for the untouched official policy.

Our experiments isolate this distinction. The paired study of Sec.~\ref{sec:necessity} fixes the student architecture, data volume, training procedure, and evaluation protocol while changing only the source of the demonstrations. In this BC-only comparison on peg, the identical student reaches $31.5\%$ success at $\sigma=5$\,mm when trained from demonstrations generated without offset access, but $88.7\%$ and $95.8\%$ when trained from privileged and relabelled demonstrations. The non-privileged student closely reproduces its teacher's collapse despite receiving the same deployable sensing. Failure analysis supports the same mechanism. Between $87\%$ and $100\%$ of its failures involve contact without insertion, and although the target remains visible, the commanded motion stays referenced to the displaced frame. Deployable sensing alone is insufficient. Robust behaviour requires supervision that encodes compensation for the latent frame offset. We make three contributions.

\begin{enumerate}
\item \textbf{A mechanistic analysis of pose-error failure under a displaced action frame.} We show that when the same estimated fixture pose enters the observation and anchors the action frame, the resulting offset is unidentifiable from proprioception before contact. A controlled teacher-swap experiment isolates offset-aware supervision, rather than deployable sensing alone, as the source of robustness.

\item \textbf{Two forms of offset-aware supervision for robust precision assembly.} Fixture-offset compensation is encoded either by a state-based reinforcement learning (RL) teacher with access to the true offset or by closed-form relabelling of clean demonstrations from the official policy. BC and DAgger distil this supervision into a student restricted to deployable sensing. Under the primary randomisation-on protocol, the student maintains at least $96.5\%$ success through $\sigma=5$\,mm, where the official policy falls to $31.8\%$ to $60.4\%$. Relabelled demonstrations recover the benefit across all three tasks, while a learned frame corrector provides an additional control on peg.

\item \textbf{Robustness that survives transfer.} Under one frozen protocol, the student outperforms six non-privileged baselines, including matched-law noise augmentation and an asymmetric critic with offset access. Deployed zero-shot on a Franka across all three tasks, it reaches $83.3\%$ pooled success at $\sigma=5$\,mm versus $34.4\%$ for the strongest state-based policy.
\end{enumerate}

\section{Related Work}
\textbf{Simulated contact-rich assembly.} Factory \cite{narang2022factory}, IndustReal \cite{tang2023industreal}, and AutoMate \cite{tang2024automate} made large-scale simulated assembly practical, increasingly robust to pose error and diverse in geometry. FORGE \cite{noseworthy2025forge} adds force-threshold conditioning and trains with pose noise hidden from the policy, exposing the coupling between observation and action frame studied here. We use its three Isaac Lab \cite{mittal2023orbit} tasks without modification. Recent FORGE-family results are not directly matched to this setting. CoRMA \cite{corma2026} uses a Marvin arm, SI-Diff \cite{sidiff2026} retrains on custom assets, FORGE-plus \cite{forgeplus2026} removes pose noise, while MATCH \cite{match2026} and a tactile study \cite{sblr2026} use Factory tasks. We therefore reproduce the unmodified FORGE gyms for comparison, preserving their task assets, action frame, reward, and success criteria.

\textbf{Privileged distillation and relabelling.} Privileged simulator state is commonly distilled into deployable sensing in locomotion \cite{lee2020learning,miki2022} and driving \cite{chen2019lbc}. DAgger \cite{ross2011dagger} addresses covariate shift, while rapid motor adaptation (RMA) \cite{kumar2021rma} distils a teacher that observes extrinsics. The work of Fuchioka et al.\ \cite{fuchioka2024}, whose teacher observes the true peg pose, is the closest precedent. Our privileged variable is instead the \emph{error} between clean and observed pose, supplied to the actor rather than only the critic \cite{pinto2018asymmetric}. Our closed-form relabelling shares SPARTN's \cite{spartn2023} algebra but perturbs the pose \emph{estimate}, leaving recorded images and forces valid without re-rendering, replayed interventions \cite{intervengen2024}, retargeting \cite{mimicgen2023}, or noise injection \cite{her2017,dart2017}. SIDO's pose-channel displacement \cite{sido2026} is the closest concurrent operation.

\textbf{Online adaptation and vision-force policies.} RMA-style adapters \cite{kumar2021rma} and CoRMA \cite{corma2026} infer context from recent history, while residual RL \cite{johannink2019} and InsertionNet \cite{spector2021} correct nominal insertion poses from vision and force. Section~\ref{sec:baselines} reimplements both strategies and tests the limitation of history-based offset inference in our setting. Vision-force fusion for insertion is established \cite{lee2019making,zhao2023act,chi2023diffusion,manipforce2025}. Our student uses ManipForce's cross-attention primitive \cite{manipforce2025} without claiming architectural novelty or attributing the robustness gain to fusion (Sec.~\ref{sec:arch}).

\begin{figure*}[t]
\centering
\providecommand{\subtxt}[1]{{\sffamily\tiny\color{black!70}#1}}%
\resizebox{\textwidth}{!}{%
\begin{tikzpicture}[ffont,
  box/.style 2 args={draw=#1, fill=#2, rounded corners=2.6pt, line width=0.8pt, inner sep=4pt, align=center, font=\sffamily\scriptsize},
  sub/.style={font=\sffamily\tiny, text=black!70},
  arr/.style={-{Stealth[length=2mm]}, line width=0.9pt},
  panel/.style={draw=sPan, dashed, fill=fPan, rounded corners=7pt, line width=0.6pt},
  ptitle/.style={font=\sffamily\bfseries\footnotesize, text=tTitle, anchor=north west}]
\draw[panel] (0,-0.35) rectangle (5.35,6.15);
\draw[panel] (5.65,-0.35) rectangle (16.45,6.15);
\draw[panel] (16.75,-0.35) rectangle (19.55,6.15);
\node[ptitle] at (0.12,6.1) {(a) Offset into the demonstrations};
\node[ptitle] at (5.77,6.1) {(b) Student: deployable sensing only};
\node[ptitle] at (16.87,6.1) {(c) Deployment};
\node[box={sPri}{fPri}, text width=2.2cm, anchor=north west] (src1) at (0.15,5.35)
  {\textbf{(i) privileged teacher T-B}\\[1pt] \subtxt{actor sees $\epsilon$, acts under}\\ \subtxt{$\sigma\!\sim\!U[0,5]$\,mm}};
\node[box={sPri}{fPri}, text width=2.2cm, anchor=north west] (src2) at (0.15,3.75)
  {\textbf{(ii) closed-form relabelling}\\[1pt] \subtxt{official T-A at $\sigma{=}0$,}\\ \subtxt{$\mathbf{o}_{0:3}{-}\epsilon$, $\mathbf{a}_{0:3}{-}\epsilon/b$}};
\node[box={sSta}{fSta}, text width=1.4cm] (demo) at (4.13,0 |- src1.east)
  {\textbf{demos in}\\ \textbf{the $\hat p$ frame}};
\node[box={sStu}{fStu}, text width=1.85cm] (bc) at (4.13,0 |- src2.east)
  {\textbf{BC $+$ one}\\ \textbf{DAgger round}\\[1pt] \subtxt{about 2000{+}1024 eps.}};
\draw[arr, sPri] (src1.east) -- (demo.west);
\draw[sPri, line width=0.9pt] (src2.east) -- ++(0.2,0) |- ($(src1.east)!0.5!(demo.west)$);
\draw[arr, sSta] (demo.south) -- (bc.north);
\begin{scope}[shift={(0.6,0.45)}, scale=1.15]
  \fill[black!10] (0,0) rectangle (3.6,0.28);
  \draw[black!55, line width=0.7pt, fill=black!22] (0.55,0.28) -- (0.55,0.95) -- (0.8,0.95) -- (0.8,0.55) -- (1.05,0.55) -- (1.05,0.95) -- (1.3,0.95) -- (1.3,0.28);
  \draw[sFor, dashed, line width=0.7pt, fill=fFor] (2.05,0.28) -- (2.05,0.95) -- (2.3,0.95) -- (2.3,0.55) -- (2.55,0.55) -- (2.55,0.95) -- (2.8,0.95) -- (2.8,0.28);
  \draw[{Stealth[length=1.5mm]}-{Stealth[length=1.5mm]}, sFor, line width=0.8pt] (0.925,1.25) -- node[above=0pt, font=\sffamily\scriptsize, text=sFor] {$\epsilon$} (2.425,1.25);
  \node[font=\sffamily\scriptsize, text=black!60, anchor=north] at (0.925,0.22) {$p^\star$};
  \node[font=\sffamily\scriptsize, text=sFor, anchor=north] at (2.425,0.22) {$\hat p$};
\end{scope}
\node[font=\sffamily\scriptsize, text=black!70, align=center, text width=4.2cm, anchor=north] at (2.675,0.3) {both put the offset $\epsilon$ into the demonstrations};
\draw[arr, sStu] (bc.east) -- (5.65,0 |- bc.east);
\node[inner sep=0, draw=sVis, line width=0.6pt] (tpimg) at (6.56,4.85) {\includegraphics[width=1.5cm]{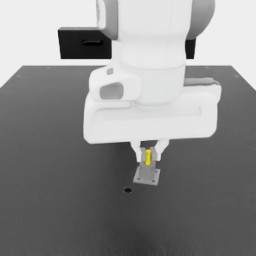}};
\node[inner sep=0, draw=sVis, line width=0.6pt] (wrimg) at (6.56,3.2) {\includegraphics[width=1.5cm]{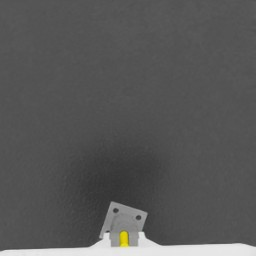}};
\node[inner sep=0, draw=sFor, line width=0.6pt] (fzimg) at (6.56,1.55) {\includegraphics[width=1.5cm]{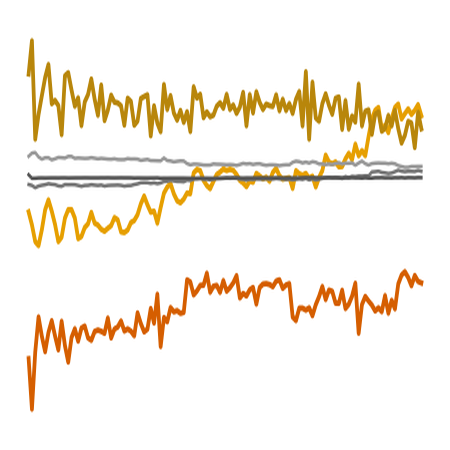}};
\node[box={sVis}{fVis}, text width=1.05cm, minimum height=1.5cm] (tp) at (8.035,4.85) {\textbf{TP RGB}\\ \subtxt{$256^2{\times}3$}};
\node[box={sVis}{fVis}, text width=1.05cm, minimum height=1.5cm] (wr) at (8.035,3.2) {\textbf{wrist RGB}\\ \subtxt{$256^2{\times}3$}};
\node[box={sFor}{fFor}, text width=1.05cm, minimum height=1.5cm] (fz) at (8.035,1.55) {\textbf{wrench}\\ \subtxt{$32{\times}6$ @ $120$\,Hz}};
\node[box={sSta}{fSta}, text width=1.05cm] (stt) at (8.035,0.285) {\textbf{state}\\ \subtxt{$24$-dim}};
\node[box={sVis}{fVis}, text width=1.35cm, minimum height=1.5cm] (r1) at (9.915,4.85) {\textbf{ResNet-18}\\ \subtxt{$8{\times}8{\times}512$}};
\node[box={sVis}{fVis}, text width=1.35cm, minimum height=1.5cm] (r2) at (9.915,3.2) {\textbf{ResNet-18}\\ \subtxt{$8{\times}8{\times}512$}};
\node[box={sFor}{fFor}, text width=1.35cm, minimum height=1.5cm] (cv) at (9.915,1.55) {\textbf{Conv1d ${\times}2$}\\ \subtxt{stride 2}};
\node[box={sSta}{fSta}, text width=1.35cm] (li) at (9.915,0.285) {\textbf{Linear}};
\draw[arr, sVis] (tp) -- (r1); \draw[arr, sVis] (wr) -- (r2); \draw[arr, sFor] (fz) -- (cv); \draw[arr, sSta] (stt) -- (li);
\node[box={sStu}{fStu}, minimum width=3.0cm, minimum height=4.8cm, anchor=north] (fus) at (12.88,5.6) {};
\node[font=\sffamily\scriptsize\bfseries, text=sStu, anchor=north] at (12.88,5.28) {bidirectional fusion};
\node[box={sVis}{fVis}, text width=1.9cm, minimum height=1.1cm, font=\sffamily\small] (vt) at (12.88,4.025) {vision\\[-1pt] \subtxt{$128$ tok}};
\node[box={sFor}{fFor}, text width=1.9cm, minimum height=1.1cm, font=\sffamily\small] (ft) at (12.88,1.55) {force\\[-1pt] \subtxt{$8$ tok}};
\foreach \k in {0,1,2,3} { \fill[sVis!70] ({12.055+\k*0.45},3.0) rectangle ++(0.3,0.24); }
\foreach \k in {0,1,2} { \fill[sFor!80] ({12.28+\k*0.45},2.35) rectangle ++(0.3,0.24); }
\foreach \k in {0,1,2} { \draw[sStu, line width=0.45pt] ({12.43+\k*0.45},2.59) -- ({12.205+\k*0.45},3.0); \draw[sStu, line width=0.45pt] ({12.43+\k*0.45},2.59) -- ({12.655+\k*0.45},3.0); }
\draw[{Stealth[length=1.6mm]}-{Stealth[length=1.6mm]}, sStu, line width=1.0pt] (13.95,2.35) -- (13.95,3.24);
\draw[arr, sVis] (r1.east) -- ++(0.25,0) |- (vt.west);
\draw[arr, sVis] (r2.east) -- ++(0.25,0) |- (vt.west);
\draw[arr, sFor] (cv.east) -- ++(0.25,0) |- (ft.west);
\node[box={sStu}{fStu}, text width=0.9cm] (ap) at (15.37,3.2) {\textbf{attn.\ pool}\\ \subtxt{$4{+}4$ tok}};
\node[box={black!70}{white}, text width=1.5cm] (mlp) at (15.37,1.55) {\textbf{MLP head}\\ \subtxt{\mbox{$9$ tok $\to a\in\mathbb{R}^{7}$}}};
\draw[arr, sStu] (fus.east) -- (ap.west);
\draw[arr, sStu] (ap.south) -- (mlp.north);
\draw[arr, sSta] (li.east) -- (15.37,0.285) -- (mlp.south);
\node[sub, anchor=north] at (12.88,0.22) {$1$ tok};
\node[inner sep=0, draw=sVis, line width=0.6pt] (realimg) at (18.15,4.3) {\includegraphics[width=2.3cm]{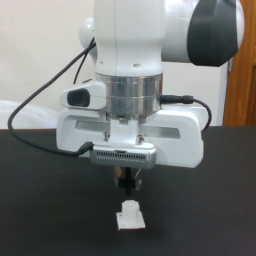}};
\node[sub, anchor=north] at (18.15,3.05) {fixed camera, Franka Panda};
\node[box={sStu}{fStu}, text width=2.2cm, anchor=north] (dep) at (18.15,2.65)
  {\textbf{same policy, same sensing}\\[1pt] \subtxt{noisy state, wrench, two RGB}\\ \subtxt{$15$\,Hz, $30$ trials per cell}\\ \subtxt{$\sigma = 0$ to $5$\,mm injected}};
\draw[arr, black!70] (mlp.east) -- node[sub, above=1pt] {$a$} (dep.west |- mlp.east);
\end{tikzpicture}}
\caption{\textbf{(a)} Two ways to encode offset compensation in demonstrations: a privileged teacher or closed-form relabelling, followed by BC and one DAgger round. \textbf{(b)} The $26.5$ M-parameter student uses only deployable state, wrench, and RGB sensing from a wrist and a third-person (TP) camera, fused into a single-shot $7$-dim action at $15$ Hz. \textbf{(c)} The same policy and sensing are deployed unchanged on the Franka.}
\label{fig:arch}
\end{figure*}
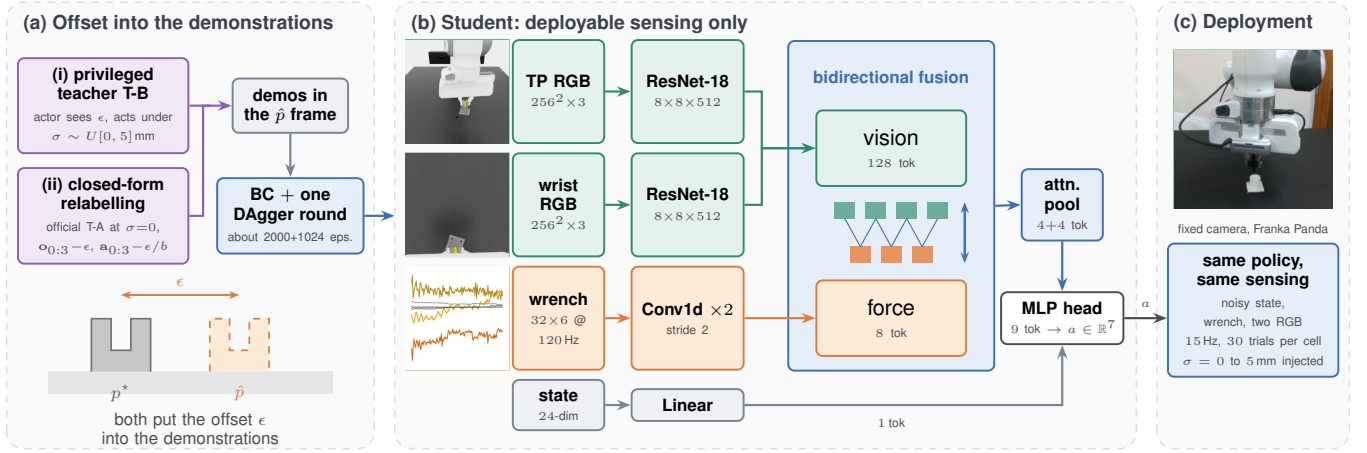

\section{The Displaced Action Frame}
\label{sec:trap}

Let $p^\star$ denote the fixture's true position and $\hat p$ the estimate. FORGE draws one offset per episode and keeps it fixed (\texttt{factory\_env.py}, lines 648 to 651):
\begin{equation}
\epsilon \sim \mathcal{N}(0,\sigma^2 I_3), \qquad \hat p = p^\star + \epsilon .
\end{equation}
The observation contains $\hat p$. By itself, this is ordinary pose-observation noise. FORGE also constructs the translational action frame from the contaminated estimate (\texttt{forge\_env.py:162}):
\begin{equation}
F_{\text{action}} = \hat p = p^\star + \epsilon .
\label{eq:trap}
\end{equation}
A translational command $\Delta p(a)$ is realised relative to $\hat p$, yielding $\hat p+\Delta p(a)$ rather than $p^\star+\Delta p(a)$. The same offset affects both the observed pose and the action frame.

Before contact, the offset is unidentifiable from proprioceptive state. Under this observation model, every pre-contact entry of the $24$-dim state is either independent of $\epsilon$ or depends on the fixture pose only through $p^\star+\epsilon$. Any pair $(p^\star,\epsilon)$ preserving this sum produces the same pose-dependent state observation. Contact is then the only state-based channel to the true geometry, but this information becomes available only after interaction begins. Vision provides geometric information before contact, while force becomes informative after contact. Vision therefore supplies geometric cues unavailable to the state alone, from which a suitably supervised policy can learn compensation.

Performance degrades once the displacement exceeds the task tolerance. Within the clearance, the nominal strategy can still succeed. Beyond it, actions correct in the believed frame are misaligned with the physical geometry. For isotropic per-axis noise, the offset magnitude has mean roughly $1.6\sigma$. Thus $\sigma=5$\,mm corresponds to about $8$\,mm mean displacement and about $6$\,mm laterally, against a $0.114$\,mm peg-hole radial clearance. The offset $\epsilon$ remains fixed per episode, making it structurally closer to a calibration bias than to sensor jitter.

FORGE's translational targets saturate at the $5$ cm bound and are re-issued every step. At $\sigma=5$\,mm, the displaced frame mainly changes the commanded direction by a few degrees, making pre-contact geometric sensing decisive (Sec.~\ref{sec:necessity}).

\textbf{Tasks.} Peg insertion uses an $8$\,mm peg and $149$ steps, gear meshing uses $299$ steps, and M16 nut threading uses $449$ steps with a yaw criterion. Control uses a $7$-dim action at $15$ Hz over $120$ Hz physics. The seventh component is a success-prediction head used only by the reward.

\section{Method}

Our method encodes offset compensation into demonstrations during training and transfers this supervision to a policy that receives only deployable sensing.

\subsection{Three tiers}

\textbf{T-A, the official baseline.} T-A is the unmodified FORGE policy with a $24$-dim noisy state, proximal policy optimisation (PPO), an MLP+LSTM actor, and a $61$-dim privileged critic. It trains for $200$ epochs under FORGE's default fixed $\sigma=1$\,mm pose noise, with three seeds per task.

\textbf{T-B, the privileged-noise teacher.} T-B matches T-A in network structure, reward, randomisation, training budget, and success criterion. It differs only in training noise and actor observation. T-B trains under a per-episode $\sigma \sim U[0,5]$\,mm law, and its actor receives the $61$-dim clean privileged state together with the $3$-dim offset $\epsilon$, for $64$ dimensions in total. T-B remains simulation-only and is never deployed. We train three seeds per task.

\textbf{The student.} The student receives only deployable inputs: the same $24$-dim noisy state available to T-A, a $32{\times}6$ window of \emph{raw, unsmoothed} wrench at the $120$ Hz physics rate, corresponding to $267$ ms of history, and two $256{\times}256$ RGB streams from a wrist camera and a fixed TP camera. T-A instead receives only a $3$-dim smoothed force signal. Two ResNet-18 trunks preserve spatial image tokens, while a strided-convolution tokeniser converts the wrench window into $8$ tokens. A bidirectional cross-attention block fuses the modalities, and an MLP head predicts the $7$-dim action in a single shot at $15$ Hz with no action chunking (Fig.~\ref{fig:arch}).

\subsection{Distillation}
\label{sec:distill}

We distil T-B rather than train the visual policy directly with RL to reduce training cost and measure how much of the teacher's pose-error robustness transfers to deployable sensing.

Demonstrations are collected by running T-B in a camera-augmented copy of the FORGE environment with per-episode $\sigma \sim U[0,5]$\,mm and retaining successful episodes. BC uses AdamW, a cosine schedule, Huber loss, and early stopping on an episode-wise held-out split.

One DAgger round follows. Because T-B is recurrent, querying it per step during a student rollout would break recurrent-state consistency. The student instead drives the rollout while the environment records T-B's $64$-dim observation as instrumentation. The complete sequence is replayed through T-B afterwards to produce labels with the correct recurrent state. Failed student trajectories are retained, and the student is retrained from scratch on the merged dataset.

\subsection{Closed-form relabelling}
\label{sec:relabel}

This route encodes offset compensation directly into the demonstrations without a privileged teacher. Because the displacement in \eqref{eq:trap} is a known frame translation, a clean-frame demonstration can be expressed in any displaced frame in closed form. We collect demonstrations from the \emph{official} T-A at $\sigma=0$, where it is $99\%$ successful, draw one $\epsilon$ per episode from the same law, and transform the recorded state and action as $\mathbf{o}_{0:3} \leftarrow \mathbf{o}_{0:3} - \epsilon$ and $\mathbf{a}_{0:3} \leftarrow \mathbf{a}_{0:3} - \epsilon/b$, where $b=0.05$ m is the translational action bound and the previous-action entries shift likewise. Images and wrench remain unchanged because the physical geometry does not change, while the shifted action reproduces the recorded motion exactly under \eqref{eq:trap}. This procedure requires neither a privileged teacher nor re-rendering under noise. Its only privileged quantity is $\epsilon$ at data-generation time. The closed form applies to translation only, since rotational pose error would require a different relabelling. Section~\ref{sec:necessity} isolates the contributions of the state and action shifts and compares this route with privileged-teacher distillation.

\section{Evaluation Protocol}
\label{sec:eval}

The evaluation protocol was frozen before any reported result was produced. Each cell contains $256$ episodes from a fixed protocol seed, so every checkpoint faces the same episode set. Actions are deterministic, and evaluation spans $\sigma\in\{0,1,2.5,5\}$\,mm $\times$ dynamics randomisation $\in\{\mathrm{on},\mathrm{off}\}$ using FORGE's native success criterion and Wilson $95\%$ intervals. Primary comparisons use three independently trained seeds, and single-seed results are marked.

The $0$ to $5$\,mm range mirrors FORGE's hardware protocol (low/medium/high: $0$ to $1$, $1$ to $2.5$, and $2.5$ to $5$\,mm), while $2.5$\,mm matches our hand-eye calibration residual. Code, the frozen protocol, and all checkpoints will be released.

We report only frozen-protocol results. Checkpoints are selected by held-out behaviour-cloning loss with early stopping, never by frozen-protocol success, and training-time statistics are not used as evidence. Unless stated otherwise, prose reports the dynamics-randomisation-on condition.

\section{Results}

\subsection{Main result}

\begin{table*}[t]
\caption{Success rate (\%) across the noise spectrum, averaged over three seeds ($n=256$/cell). Cells report dynamics randomisation on/off. T-A+ retrains T-A at $\sigma=5$\,mm, and T-B is the privileged oracle.}
\label{tab:main}
\centering
\small
\begin{tabular*}{\textwidth}{@{\extracolsep{\fill}}llcccc@{}}
\toprule
task & model & $0$\,mm & $1$\,mm & $2.5$\,mm & $5$\,mm \\
\midrule
\multirow{4}{*}{peg}
 & T-A official baseline        & 98.7 / 98.2 & 97.3 / 94.7 & 73.7 / 74.5 & 31.8 / 33.3 \\
 & T-A+ noise-augmented         & 84.1 / 77.3 & 78.0 / 75.1 & 58.7 / 56.2 & 29.9 / 25.8 \\
 & T-B privileged (oracle)      & 99.3 / 96.6 & 99.3 / 96.7 & 99.7 / 97.5 & 99.3 / 96.6 \\
 & \textbf{student (ours)}      & \textbf{98.3 / 96.2} & \textbf{98.6 / 96.2} & \textbf{98.3 / 95.4} & \textbf{96.6 / 91.8} \\
\midrule
\multirow{4}{*}{gear}
 & T-A official baseline        & 98.6 / 100.0 & 97.5 / 99.2 & 86.9 / 90.6 & 49.9 / 53.8 \\
 & T-A+ noise-augmented         & 94.9 / 99.7 & 94.7 / 99.7 & 92.1 / 99.3 & 77.9 / 90.0 \\
 & T-B privileged (oracle)      & 98.7 / 99.7 & 99.0 / 100.0 & 98.2 / 99.7 & 98.2 / 98.8 \\
 & \textbf{student (ours)}      & \textbf{98.8 / 98.4} & \textbf{99.2 / 97.8} & \textbf{99.3 / 96.5} & \textbf{98.2 / 96.5} \\
\midrule
\multirow{4}{*}{nut}
 & T-A official baseline        & 97.4 / 97.8 & 96.1 / 96.9 & 87.4 / 87.9 & 60.4 / 64.3 \\
 & T-A+ noise-augmented         & 89.7 / 88.4 & 88.8 / 88.3 & 83.7 / 85.2 & 73.0 / 71.2 \\
 & T-B privileged (oracle)      & 99.2 / 99.7 & 99.5 / 99.9 & 99.9 / 99.6 & 99.2 / 99.9 \\
 & \textbf{student (ours)}      & \textbf{99.1 / 98.8} & \textbf{99.0 / 98.8} & \textbf{99.0 / 98.6} & \textbf{96.5 / 95.1} \\
\bottomrule
\end{tabular*}
\end{table*}

Across the three tasks (Table~\ref{tab:main}), the student matches the official baseline at $0$\,mm and outperforms it at every nonzero noise level, with the largest gains at higher pose error. At $5$\,mm, the improvement reaches $+64.8$ points on peg, $+48.3$ on gear, and $+36.1$ on nut. When measured as the fraction of baseline failures eliminated, the student removes $92.1\%$ to $94.7\%$ of failures at $2.5$\,mm and $91.1\%$ to $96.4\%$ at $5$\,mm. The recovered oracle gap, defined as $(\text{student}-\text{T-A})/(\text{T-B}-\text{T-A})$, is $96.0\%$ on peg, $100\%$ on gear, and $93.0\%$ on nut at $5$\,mm.

\textbf{Noise-augmented state baseline.} T-A was not trained at these noise levels, so T-A+ retrains the same policy at $\sigma=5$\,mm with all other settings unchanged. Its response varies across tasks. Gear improves from $49.9\%$ to $77.9\%$, nut improves from $60.4\%$ to $73.0\%$, while \textbf{peg does not adapt to the noisy regime}, reaching only $29.9\%$ at $5$\,mm and dropping from $98.7\%$ to $84.1\%$ at $0$\,mm. Under \eqref{eq:trap}, the reward is evaluated in the true frame while the action is executed in the displaced frame, causing identical actions to receive different returns for factors hidden from the policy and thereby degrading credit assignment. Training T-A under the student's $U[0,5]$\,mm law reaches $31.6/74.2/67.6\%$ at $5$\,mm on peg/gear/nut. Letting only the critic see $\epsilon$ changes little ($35.6/79.7/71.9\%$, Table~\ref{tab:baselines}): the offset still does not reach the actor's supervision.

\subsection{Where the student's robustness comes from}
\label{sec:necessity}

This ablation tests whether the student's advantage is explained by access to
cameras or by the supervision contained in the demonstrations. Three arms (Table~\ref{tab:necessity}) differ
only in how the demonstrations were made (same environment, noise law, success
filter, budget of about $2000$ episodes, trainer and protocol): one distils
from privileged T-B, one (the ``weak'' arm) from the official non-privileged
T-A acting under noise, and one from T-A at $\sigma{=}0$ with the closed-form relabelling of Sec.~\ref{sec:relabel}. All are BC-only.

The first three rows of Table~\ref{tab:necessity} are the same student trained on demonstrations from three sources: T-A run under the test
noise (it cannot see $\epsilon$, so its demonstrations contain no compensation),
the privileged teacher T-B (it sees $\epsilon$, so they do), and T-A run at
$\sigma{=}0$ with the relabelling applied afterwards. The two control rows take
the same clean demonstrations and either apply no shift or shift only the state
pose entry. The last two rows are the teachers themselves under the noise, since
a student cannot be expected to exceed its own teacher.

\begin{table}[t]
\caption{Matched peg ablation ($n=256$/cell). The first three rows differ only in demonstration construction and report three-seed means $\pm$ sd. Controls and $+$DAgger use seed~0.}
\label{tab:necessity}
\centering
\scriptsize
\setlength{\tabcolsep}{3pt}
\begin{tabular*}{\columnwidth}{@{\extracolsep{\fill}}lcccc@{}}
\toprule
demonstrations & $0$\,mm & $2.5$\,mm & $5$\,mm & $10$\,mm \\
\midrule
T-A under noise (3 seeds) & 98.6$\pm$0.8 & 70.6$\pm$1.9 & \textbf{31.5$\pm$2.2} & 8.7$\pm$0.5 \\
priv.\ T-B (3 seeds) & 95.6$\pm$3.3 & 93.9$\pm$6.3 & 88.7$\pm$7.1 & 70.7$\pm$9.7 \\
clean, relabelled (3 seeds) & 98.0$\pm$0.8 & 96.5$\pm$1.8 & \textbf{95.8$\pm$1.0} & 84.2$\pm$2.6 \\
\midrule
clean, no shift (s0) & 98.0 & 63.7 & 26.2 & 7.8 \\
clean, state shift only (s0) & 98.4 & 98.8 & 94.1 & 66.8 \\
clean, relabelled $+$ DAgger (s0) & 98.8 & 99.6 & 100.0 & 94.5 \\
\midrule
\emph{T-A teacher itself}  & 98.8 & 77.7 & 35.2 & 12.5 \\
\emph{T-B teacher itself}  & 99.6 & 100.0 & 99.2 & 97.7 \\
\bottomrule
\end{tabular*}
\end{table}

At $\sigma=5$\,mm, the weak arm trails the privileged-teacher and relabelled arms by $57$ and $64$ points, respectively, across three seeds. Its cross-seed spread is only $2.2$ points, and its seed-wise Wilson intervals are disjoint from those of both other arms. The weak student also closely tracks its non-privileged teacher's degradation, identifying demonstration construction as the dominant factor in this matched BC comparison. The same pattern appears in the single-seed gear and nut controls, where the non-privileged students reach $56.6\%$ and $48.8\%$ at $\sigma=5$\,mm, versus $98.2\%$ and $96.5\%$ for the full students.

\textbf{Effect of closed-form relabelling.} On peg, the identical student trained on $2028$ relabelled clean demonstrations matches or exceeds the BC-only privileged-teacher arm and remains within $2$ points of the full BC+DAgger student through $\sigma=5$\,mm. On gear and nut, the BC-only relabelled students reach $95.1\%$ and $94.5\%$ at $\sigma=5$\,mm, compared with $98.2\%$ and $96.5\%$ for the full students. Adding the same DAgger round raises the relabelled peg arm to $100.0\%$ at $\sigma=5$\,mm and $94.5\%$ at $10$\,mm. For the translational offset studied here, relabelling therefore provides nearly the same offset-aware supervision as the privileged teacher, while the teacher route remains applicable when closed-form relabelling is unavailable.

\textbf{Components of the relabelling.} With no shift, the student reproduces the weak arm's collapse. Shifting only the state-pose entry yields $94.1\%$ at $\sigma=5$\,mm but $66.8\%$ at $10$\,mm. The state shift explains most of the in-range recovery, while the action shift becomes more important beyond the training range. The weak arm's $75\%$ acceptance rate favours easier offsets, so success filtering cannot explain the gap.

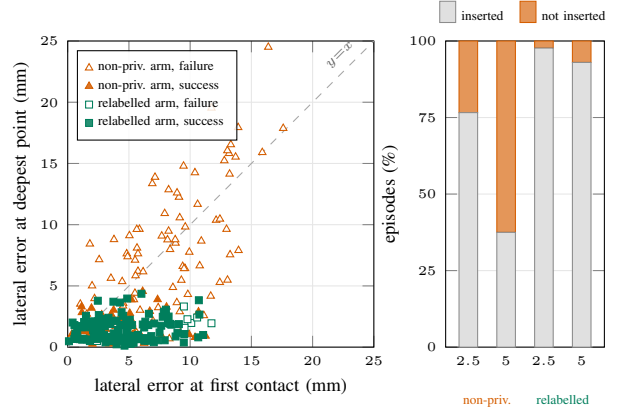
\begin{figure}[t]
\centering
\begin{tikzpicture}[font=\scriptsize]
\begin{axis}[name=sc, width=4.05cm, height=4.05cm, scale only axis,
  xmin=0, xmax=25, ymin=0, ymax=25, xtick={0,5,10,15,20,25}, ytick={0,5,10,15,20,25},
  xlabel={lateral error at first contact (mm)}, ylabel={lateral error at deepest point (mm)},
  xlabel shift=-2pt, ylabel shift=-3pt, tick label style={font=\tiny}, grid=major, grid style={gray!20},
  legend style={at={(0.03,0.97)}, anchor=north west, font=\tiny, fill=white, fill opacity=0.85, text opacity=1,
                row sep=-2pt, /tikz/every even column/.append style={column sep=2pt}}, legend cell align=left]
\addplot[gray!70, dashed, thin, domain=0:25, samples=2, forget plot] {x};
\node[anchor=south east, font=\tiny, text=gray!80!black, rotate=45] at (axis cs:24.6,24.6) {$y{=}x$};
\addplot[only marks, mark=triangle*, mark size=1.6pt, cbVerm, mark options={fill=white}] coordinates {(12.97,9.63) (7.39,3.23) (8.15,8.77) (12.12,10.37) (1.56,2.70) (9.58,6.43) (2.57,7.15) (8.91,3.64) (5.91,3.02) (13.26,7.55) (4.53,3.59) (4.45,5.36) (5.66,8.10) (9.28,5.47) (1.03,3.52) (9.81,4.31) (13.18,15.82) (3.91,1.68) (8.30,0.69) (9.46,14.79) (7.92,10.92) (4.52,3.37) (10.39,14.25) (4.82,7.63) (5.74,6.20) (5.54,5.84) (5.50,7.11) (17.60,17.87) (12.42,5.29) (3.53,5.63) (15.89,15.90) (11.68,4.17) (5.49,4.41) (9.93,7.76) (13.22,14.14) (10.61,11.67) (11.74,19.53) (5.26,1.91) (5.04,9.11) (5.74,7.63) (9.15,10.57) (4.77,1.06) (8.81,8.50) (2.19,3.97) (8.59,4.88) (9.09,12.25) (12.43,10.46) (8.24,12.86) (9.15,2.90) (5.86,9.63) (13.32,16.54) (5.64,9.62) (9.39,6.60) (16.38,24.50) (8.36,7.98) (6.16,6.16) (9.60,9.82) (13.71,15.53) (13.92,17.94) (7.68,9.08) (8.93,12.60) (13.95,7.90) (6.97,2.13) (8.78,8.78) (10.78,6.65) (10.91,8.67) (5.41,3.88) (5.62,1.19) (1.99,5.01) (9.50,6.42) (7.13,13.88) (12.79,15.23) (8.24,9.48) (13.05,5.48) (4.89,7.39) (3.78,8.79) (6.91,13.36) (13.04,16.05) (11.12,2.58) (1.81,8.43)};
\addlegendentry{non-priv.\ arm, failure}
\addplot[only marks, mark=triangle*, mark size=1.6pt, cbVerm] coordinates {(6.24,0.33) (0.89,0.98) (3.29,1.89) (1.36,2.10) (3.55,2.69) (5.20,3.88) (5.07,2.77) (6.53,0.68) (6.23,0.55) (0.34,1.06) (4.93,1.09) (1.16,2.83) (7.34,3.89) (2.57,1.29) (2.77,1.23) (0.69,1.37) (10.68,2.88) (3.75,3.03) (1.23,1.28) (9.91,1.04) (1.62,1.28) (2.50,2.49) (9.32,1.40) (4.33,2.34) (4.36,0.57) (4.53,0.48) (4.20,2.31) (5.12,2.34) (3.07,1.99) (2.02,0.26) (5.95,0.99) (3.12,1.47) (2.78,2.64) (6.95,1.88) (1.70,0.98) (5.43,1.10) (6.70,2.93) (2.87,1.95) (5.28,1.02) (6.12,4.56) (3.41,0.27) (8.07,3.28) (10.40,0.80) (1.13,3.30) (1.94,3.19) (4.41,0.67) (11.26,0.91) (7.64,0.96)};
\addlegendentry{non-priv.\ arm, success}
\addplot[only marks, mark=square*, mark size=1.3pt, cbGreen!80!black, mark options={fill=white}] coordinates {(4.95,2.17) (0.60,0.88) (10.09,1.97) (10.57,2.42) (9.54,1.58) (9.49,3.32) (11.74,1.95) (2.39,0.50) (9.78,2.28)};
\addlegendentry{relabelled arm, failure}
\addplot[only marks, mark=square*, mark size=1.3pt, cbGreen!80!black] coordinates {(5.45,0.89) (5.74,1.00) (2.00,1.97) (5.32,0.93) (7.50,1.20) (6.34,2.08) (2.97,1.79) (10.66,0.80) (2.72,1.99) (8.02,3.06) (1.84,1.89) (3.98,0.93) (1.12,1.63) (11.06,0.99) (4.38,0.77) (4.53,2.56) (1.81,0.49) (3.06,2.31) (8.19,1.96) (4.37,1.82) (7.07,1.93) (0.55,1.80) (2.72,1.13) (1.18,0.61) (6.02,4.36) (0.74,0.67) (6.63,2.72) (4.11,1.41) (8.24,2.50) (6.55,0.52) (2.49,2.11) (7.03,0.40) (4.86,3.97) (4.89,1.73) (2.73,0.72) (6.34,0.74) (3.73,3.71) (1.57,0.63) (9.48,0.38) (8.96,1.17) (2.15,1.64) (3.66,0.49) (0.91,1.72) (7.85,2.89) (4.29,1.79) (8.37,1.90) (3.98,0.64) (6.13,1.85) (5.38,0.35) (3.46,1.88) (8.46,1.83) (4.26,3.64) (4.27,0.47) (4.10,1.22) (2.13,0.78) (2.15,2.11) (3.65,2.15) (3.62,0.84) (1.34,2.05) (6.28,0.52) (7.64,1.75) (0.78,1.19) (10.71,3.83) (4.40,2.85) (0.54,0.67) (5.26,1.80) (5.76,0.67) (4.91,0.80) (4.87,1.36) (4.85,0.63) (3.99,2.16) (3.01,0.77) (7.92,0.56) (9.08,1.88) (1.45,1.62) (2.39,0.76) (3.55,0.58) (2.52,1.39) (4.83,0.61) (3.09,0.39) (0.99,0.65) (4.78,2.00) (2.84,0.94) (3.75,1.12) (5.46,0.65) (3.41,0.74) (2.06,1.39) (2.54,2.62) (6.82,0.90) (0.63,1.10) (4.01,0.67) (7.76,2.48) (6.58,1.80) (0.10,0.48) (4.64,0.12) (1.49,2.56) (10.75,2.66) (5.02,0.44) (2.75,1.70) (2.22,0.94) (2.97,0.98) (7.32,0.56) (5.07,2.24) (2.46,3.79) (8.81,0.87) (6.25,0.59) (2.41,0.67) (3.99,0.23) (4.78,1.35) (0.40,2.01) (9.52,1.07) (3.33,2.40) (3.46,1.92) (5.65,0.65) (3.02,1.70) (3.28,0.70) (5.14,0.27) (6.25,1.60) (1.30,0.69)};
\addlegendentry{relabelled arm, success}
\end{axis}
\begin{axis}[at={(sc.east)}, anchor=west, xshift=0.95cm, width=2.1cm, height=4.05cm, scale only axis, clip=false,
  ybar stacked, bar width=7pt, xmin=-0.6, xmax=3.6, ymin=0, ymax=100, ytick={0,25,50,75,100},
  ylabel={episodes (\%)}, ylabel shift=-3pt, tick label style={font=\tiny},
  xtick={0,1,2,3}, xticklabels={$2.5$,$5$,$2.5$,$5$},
  ymajorgrids, grid style={gray!20},
  legend style={at={(0.5,1.02)}, anchor=south, font=\tiny, draw=none, fill=none, legend columns=2,
                /tikz/every even column/.append style={column sep=4pt}}, legend cell align=left]
\addplot[fill=black!12, draw=black!45] coordinates {(0,76.6) (1,37.5) (2,97.7) (3,93.0)};
\addlegendentry{inserted}
\addplot[fill=cbVerm!65, draw=cbVerm] coordinates {(0,23.4) (1,62.5) (2,2.3) (3,7.0)};
\addlegendentry{not inserted}
\node[font=\tiny, text=cbVerm, anchor=north] at (axis cs:0.5,-12.5) {non-priv.};
\node[font=\tiny, text=cbGreen!80!black, anchor=north] at (axis cs:2.5,-12.5) {relabelled};
\end{axis}
\end{tikzpicture}
\caption{Where the weak arm presses (peg, $128$ paired episodes per arm and noise level, identical offsets).
\textbf{Left:} at $\sigma=5$\,mm the non-privileged arm first touches a median $6.2$\,mm off the axis and its
failures stay on the diagonal (median $7.7$\,mm at depth), whereas the relabelled arm touches at $4.3$\,mm and is
on-axis at depth ($1.4$\,mm). \textbf{Right:} outcomes at $\sigma=2.5$ and $5$\,mm. All but one of the $122$
non-inserted episodes never entered the hole.}
\label{fig:mech}
\end{figure}

\textbf{Failure-mode analysis.} Across noise levels, $87\%$ to $100\%$ of weak-arm failures are \textsc{no\_insert}. In $128$ paired episodes at $\sigma=5$\,mm (Fig.~\ref{fig:mech}), its first contact is a median $6.2$\,mm off-axis and remains displaced at depth ($7.7$\,mm), whereas the relabelled arm reaches $4.3$\,mm at first contact and $1.4$\,mm at depth. Both use the same visual sensing, but only offset-aware supervision consistently redirects motion toward the true target.

\begin{figure*}[t]
\centering
\begin{tikzpicture}[font=\scriptsize]
\begin{groupplot}[
  group style={group size=4 by 1, horizontal sep=0.5cm,
               ylabels at=edge left, yticklabels at=edge left},
  width=0.30\textwidth, height=4.3cm,
  xmin=-0.6, xmax=20.6, ymin=0, ymax=103,
  xtick={0,5,10,15,20}, ytick={0,25,50,75,100},
  xlabel={$\sigma$ (mm)}, ylabel={success rate (\%)},
  xlabel shift=-2pt, ylabel shift=-3pt,
  grid=major, grid style={gray!20},
  tick label style={font=\tiny},
  every axis title/.append style={font=\scriptsize, yshift=-3pt},
]
\nextgroupplot[title={peg insertion},
  legend to name=robustlegend, legend columns=4,
  legend style={font=\tiny, /tikz/every even column/.append style={column sep=5pt}}]
\draw[dashed, gray] (axis cs:5,0) -- (axis cs:5,103);
\addplot[black!75, dashed, line width=0.9pt, mark=square*, mark size=1pt] coordinates
  {(0,99.3)(1,99.3)(2.5,99.7)(5,99.3)(7.5,99.2)(10,97.7)(15,89.5)(20,77.7)};
\addlegendentry{T-B (oracle)}
\addplot[cbBlue, line width=1.25pt, mark=*, mark size=1.2pt] coordinates
  {(0,98.3)(1,98.6)(2.5,98.3)(5,96.6)(7.5,94.5)(10,88.8)(15,74.2)(20,55.5)};
\addlegendentry{student (ours)}
\addplot[cbVerm, line width=0.9pt, mark=triangle*, mark size=1.4pt] coordinates
  {(0,98.7)(1,97.3)(2.5,73.7)(5,31.8)(7.5,20.7)(10,12.5)(15,3.9)(20,1.6)};
\addlegendentry{T-A (official)}
\addplot[cbOrange, line width=0.9pt, mark=diamond*, mark size=1.4pt] coordinates
  {(0,84.1)(1,78.0)(2.5,58.7)(5,29.9)(7.5,21.5)(10,11.7)(15,5.9)(20,3.1)};
\addlegendentry{T-A+ (noise-aug.)}
\addlegendimage{cbBlue, line width=1.25pt, mark=square*, mark size=1.2pt}
\addlegendentry{from priv.\ T-B (seed 0)}
\addlegendimage{cbVerm, line width=1.25pt, mark=diamond*, mark size=1.4pt}
\addlegendentry{from non-priv.\ T-A}
\addlegendimage{cbVerm!60, dashed, line width=0.8pt}
\addlegendentry{T-A itself (its teacher)}
\addlegendimage{cbGreen!80!black, line width=1.1pt, mark=square*, mark size=1.2pt}
\addlegendentry{clean T-A, relabelled}
\addlegendimage{cbGreen!80!black, densely dashdotted, line width=1.25pt, mark=diamond*, mark size=1.6pt}
\addlegendentry{relabelled $+$ DAgger (seed 0)}
\addlegendimage{cbGreen!80!black, dashed, line width=0.8pt}
\addlegendentry{state shift only}
\addlegendimage{black!60, dotted, line width=0.9pt}
\addlegendentry{no shift}
\nextgroupplot[title={gear meshing}]
\draw[dashed, gray] (axis cs:5,0) -- (axis cs:5,103);
\addplot[black!75, dashed, line width=0.9pt, mark=square*, mark size=1pt] coordinates
  {(0,98.7)(1,99.0)(2.5,98.2)(5,98.2)(7.5,95.3)(10,92.6)(15,87.5)(20,74.6)};
\addplot[cbBlue, line width=1.25pt, mark=*, mark size=1.2pt] coordinates
  {(0,98.8)(1,99.2)(2.5,99.3)(5,98.2)(7.5,95.4)(10,89.1)(15,71.9)(20,55.5)};
\addplot[cbVerm, line width=0.9pt, mark=triangle*, mark size=1.4pt] coordinates
  {(0,98.6)(1,97.5)(2.5,86.9)(5,49.9)(7.5,24.2)(10,11.7)(15,5.9)(20,3.5)};
\addplot[cbOrange, line width=0.9pt, mark=diamond*, mark size=1.4pt] coordinates
  {(0,94.9)(1,94.7)(2.5,92.1)(5,77.9)(7.5,60.9)(10,42.6)(15,17.2)(20,9.8)};
\nextgroupplot[title={nut threading}]
\draw[dashed, gray] (axis cs:5,0) -- (axis cs:5,103);
\addplot[black!75, dashed, line width=0.9pt, mark=square*, mark size=1pt] coordinates
  {(0,99.2)(1,99.5)(2.5,99.9)(5,99.2)(7.5,98.8)(10,98.4)(15,93.0)(20,85.2)};
\addplot[cbBlue, line width=1.25pt, mark=*, mark size=1.2pt] coordinates
  {(0,99.1)(1,99.0)(2.5,99.0)(5,96.5)(7.5,92.3)(10,84.4)(15,64.6)(20,47.4)};
\addplot[cbVerm, line width=0.9pt, mark=triangle*, mark size=1.4pt] coordinates
  {(0,97.4)(1,96.1)(2.5,87.4)(5,60.4)(7.5,40.2)(10,25.4)(15,12.1)(20,9.4)};
\addplot[cbOrange, line width=0.9pt, mark=diamond*, mark size=1.4pt] coordinates
  {(0,89.7)(1,88.8)(2.5,83.7)(5,73.0)(7.5,60.9)(10,45.7)(15,19.5)(20,10.9)};
\nextgroupplot[title={peg: swap the teacher}, xmin=-0.4, xmax=10.4,
               xtick={0,2.5,5,7.5,10}]
\draw[dashed, gray] (axis cs:5,0) -- (axis cs:5,103);
\addplot[cbBlue, line width=1.25pt, mark=square*, mark size=1.2pt] coordinates
  {(0,95.7)(1,97.7)(2.5,98.4)(5,95.3)(7.5,89.8)(10,81.3)};
\addplot[cbVerm, line width=1.25pt, mark=diamond*, mark size=1.4pt] coordinates
  {(0,98.6)(1,96.2)(2.5,70.6)(5,31.5)(7.5,15.9)(10,8.7)};
\addplot[cbVerm!60, dashed, line width=0.8pt, mark=none] coordinates
  {(0,98.8)(1,97.3)(2.5,77.7)(5,35.2)(7.5,20.7)(10,12.5)};
\addplot[cbGreen!80!black, line width=1.1pt, mark=square*, mark size=1.2pt] coordinates
  {(0,98.0)(1,97.8)(2.5,96.5)(5,95.8)(7.5,92.8)(10,84.2)};
\addplot[cbGreen!80!black, densely dashdotted, line width=1.25pt, mark=diamond*, mark size=1.6pt] coordinates
  {(0,98.8)(2.5,99.6)(5,100.0)(10,94.5)};
\addplot[cbGreen!80!black, dashed, line width=0.8pt, mark=none] coordinates
  {(0,98.44)(1,97.27)(2.5,98.83)(5,94.14)(7.5,82.81)(10,66.80)};
\addplot[black!60, dotted, line width=0.9pt, mark=none] coordinates
  {(0,98.05)(2.5,63.67)(5,26.17)(10,7.81)};
\end{groupplot}
\node[anchor=south] at ($(group c1r1.north west)!0.5!(group c4r1.north east)+(0,0.42cm)$)
  {\pgfplotslegendfromname{robustlegend}};
\end{tikzpicture}
\caption{Success across the noise spectrum ($n=256$/cell, three-seed means within the training range). Beyond the dashed $5$\,mm training boundary, T-A, T-A+, and T-B are single-seed, while the student is single-seed on peg beyond $10$\,mm. \textbf{Right:} the teacher swap of Table~\ref{tab:necessity}, with all arms BC-only except the $+$DAgger curve (three-seed means). The non-privileged arm tracks its teacher's degradation, whereas relabelled demonstrations match the privileged-teacher result. The state shift explains most of the recovery, no shift reproduces the collapse, and DAgger brings the relabelled peg arm to the oracle level.}
\label{fig:curves}
\end{figure*}
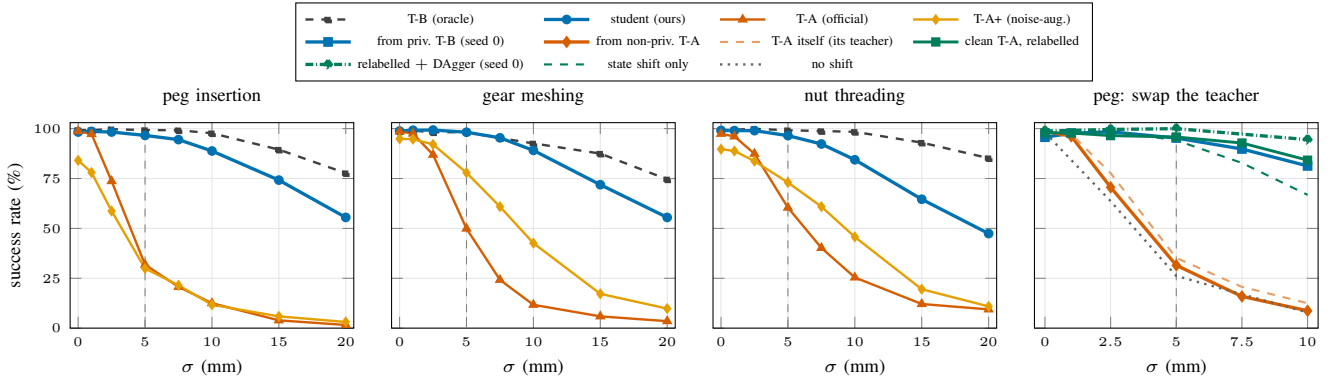

\begin{table}[t]
\caption{Success rate (\%) at $2.5/5$\,mm with dynamics randomisation on under one frozen protocol. Our method, T-A, and T-A+ use three seeds. External baselines are single-seed re-implementations: SRSA is outside its released skill library, and C3 is an RMA-style adapter rather than CoRMA itself. The matched-$U[0,5]$ and $\epsilon$-critic T-A variants are also single-seed controls.}
\label{tab:baselines}
\centering
\scriptsize
\setlength{\tabcolsep}{2pt}
\begin{tabular*}{\columnwidth}{@{\extracolsep{\fill}}llccc@{}}
\toprule
method & family & peg & gear & nut \\
\midrule
T-B & privileged oracle (upper bound) & 99.7/99.3 & 98.2/98.2 & 99.9/99.2 \\
\midrule
SRSA \cite{tang2025srsa} & skill retrieval & \phantom{0}3.1/\phantom{0}2.3 & 13.7/13.3 & 22.3/17.2 \\
C3 & RMA-style adaptation \cite{corma2026} & 59.4/30.1 & 69.5/46.5 & 76.2/54.3 \\
ARCH-INSERT \cite{arch2025} & hierarchical RL & 64.5/29.7 & 89.5/53.9 & 88.7/60.5 \\
T-A ($\epsilon$ critic) & asymmetric critic \cite{pinto2018asymmetric} & 73.1/35.6 & 96.5/79.7 & 91.0/71.9 \\
T-A ($U$ law) & noise-aug., $U[0,5]$ law & 74.6/31.6 & 91.0/74.2 & 91.8/67.6 \\
T-A+ & noise-aug.\ RL          & 58.7/29.9 & 92.1/77.9 & 83.7/73.0 \\
T-A \cite{noseworthy2025forge} & official state RL & 73.7/31.8 & 86.9/49.9 & 87.4/60.4 \\
\midrule
\textbf{student (ours)} & vision+force distil. & \textbf{98.3/96.6} & \textbf{99.3/98.2} & \textbf{99.0/96.5} \\
\bottomrule
\end{tabular*}
\end{table}

\subsection{Comparison with non-privileged baselines and a frame corrector}
\label{sec:baselines}

The four T-A variants share the same policy architecture and reward, differing in training noise and privileged information available to the critic. T-A is the official policy trained with $\sigma=1$\,mm fixture noise. T-A+ retrains it at the fixed $\sigma=5$\,mm evaluation setting, while T-A ($U$ law) uses the student's per-episode $\sigma\sim U[0,5]$\,mm law. T-A ($\epsilon$ critic) additionally exposes $\epsilon$ to the critic but not the actor. None gives the actor direct access to the offset, and all remain substantially below the student at large pose error (Table~\ref{tab:baselines}).

\emph{ARCH-INSERT} is the insertion primitive of the hierarchical assembly method in \cite{arch2025}, reimplemented because the complete system is not portable. Correcting missing velocity feedback and a rotation-scale mismatch, both of which otherwise yield zero success, produces performance that closely tracks T-A.

\emph{C3} is an RMA-style adapter inspired by CoRMA \cite{corma2026}, rather than a faithful reproduction because its 6-D contact context is unpublished. It regresses the pose offset from deployable observation history, with a validation root-mean-square error (RMSE) of $1.08$ to $1.15$\,mm. At $\sigma=5$\,mm, the offset-conditioned teacher reaches $93.0/98.4/97.7\%$ with the true offset, $30.1/46.5/54.3\%$ with the inferred offset, and $19.9/47.3/62.1\%$ with zero offset on peg/gear/nut. Online inference provides no consistent gain over zero offset ($+10/-1/-8$ points). Success depends strongly on pre-contact alignment, where force history provides little geometric information and the adapter predicts offsets near zero.

\emph{SRSA} \cite{tang2025srsa} is evaluated outside the coverage of its released retrieval library, which contains AutoMate geometries. Its $2\%$ to $3\%$ success therefore does not characterise the method. The port itself is functional, as PPO trained from scratch in the same environment reaches $66.4\%$, so we treat this comparison as inconclusive.

\textbf{Modular frame correction.}
As a positive control on peg, a ResNet-18 regressor estimates $\epsilon$ from the cameras and corrects both the observation and action frame for the \emph{unmodified} official policy. Trained on the same $2028$ clean demonstrations with synthetic offsets, it achieves $0.39$\,mm validation RMSE. Applied during the first $10$ steps, it reduces the mean residual offset at $\sigma=5$\,mm from $8.3$ to $1.0$\,mm and raises T-A from $32.4\%$ to $98.4\%$, while reaching $99.2\%$ at $\sigma=0$ and $2.5$\,mm ($n=256$). The policy itself is unchanged, showing that recovering and correcting the displaced frame is sufficient to restore performance on peg.

\begin{table}[t]
\caption{Modality ablation, retrained per condition (\%, $0/2.5/5$\,mm, dynamics randomisation on). Cells marked $^\dagger$ report three seeds (no-force SD: peg $10$ to $15$, gear $\le 2.6$, nut $7$ to $9$). All others are single-seed and noted in Sec.~\ref{sec:limits}.}
\label{tab:modality}
\centering
\scriptsize
\setlength{\tabcolsep}{2pt}
\begin{tabular*}{\columnwidth}{@{\extracolsep{\fill}}lcccc@{}}
\toprule
task & full & no force & no vision & state only \\
\midrule
peg  & 98.3/98.3/96.6$^\dagger$ & 86.3/82.2/78.1$^\dagger$ & 47.5$^\dagger$/50.0/30.7$^\dagger$ & 35.5/30.5/19.1 \\
gear & 98.8/99.3/98.2$^\dagger$ & 98.8/98.0/98.2$^\dagger$ & 91.8/82.0/63.7 & 68.4/49.2/34.8 \\
nut  & 99.1/99.0/96.5$^\dagger$ & 92.2/91.2/91.7$^\dagger$ & 76.6/68.4/52.0 & 49.2/50.0/32.0 \\
\bottomrule
\end{tabular*}
\end{table}

\subsection{What each modality does}
\label{sec:modality}

Each modality is removed in both \emph{training and test} and the student retrained (Table~\ref{tab:modality}), so the ablation measures its contribution rather than the dependence of a co-adapted network.

Vision provides the largest contribution. Removing it costs $34$ to $66$ points at $\sigma=5$\,mm, and every vision-free variant degrades sharply with pose noise. Part of this gap reflects imitation difficulty \cite{weihs2021imitationgap,swamy2022sequence}: vision-free students already underperform at $\sigma=0$ ($35\%$ to $92\%$, versus $97\%$ to $99\%$ for T-A with the same state), so this ablation upper-bounds vision's contribution.

Across three seeds, removing force costs $0$ to $1$ point on gear and $5$ to $8$ on nut, but $12/16/18$ points on peg at $\sigma=0/2.5/5$\,mm, with cross-seed spreads of $10$ to $15$ points. No-force policies press $8\%$ to $11\%$ harder and state-only policies $31\%$ to $39\%$ harder. FORGE's reported nut drop from $69\%$ to $40\%$ without force does not reproduce here: our vision-equipped student retains $91\%$ to $92\%$. With vision, force contributes little on gear, more on nut, and substantially on peg, the tightest-clearance task.

\begin{figure*}[t]
\centering
{\scriptsize\sffamily
\setlength{\tabcolsep}{1.5pt}
\newcommand{\sceneimg}[1]{\includegraphics[width=0.148\textwidth]{render/scene_#1.png}}
\begin{tabular}{@{}c@{\hspace{3pt}}cc@{\hspace{7pt}}cc@{\hspace{7pt}}cc@{}}
 & \multicolumn{2}{c}{peg insertion} & \multicolumn{2}{c}{gear meshing} & \multicolumn{2}{c}{nut threading} \\
 & initial & goal & initial & goal & initial & goal \\
\rotatebox{90}{\makebox[0.148\textwidth]{simulation}} & \sceneimg{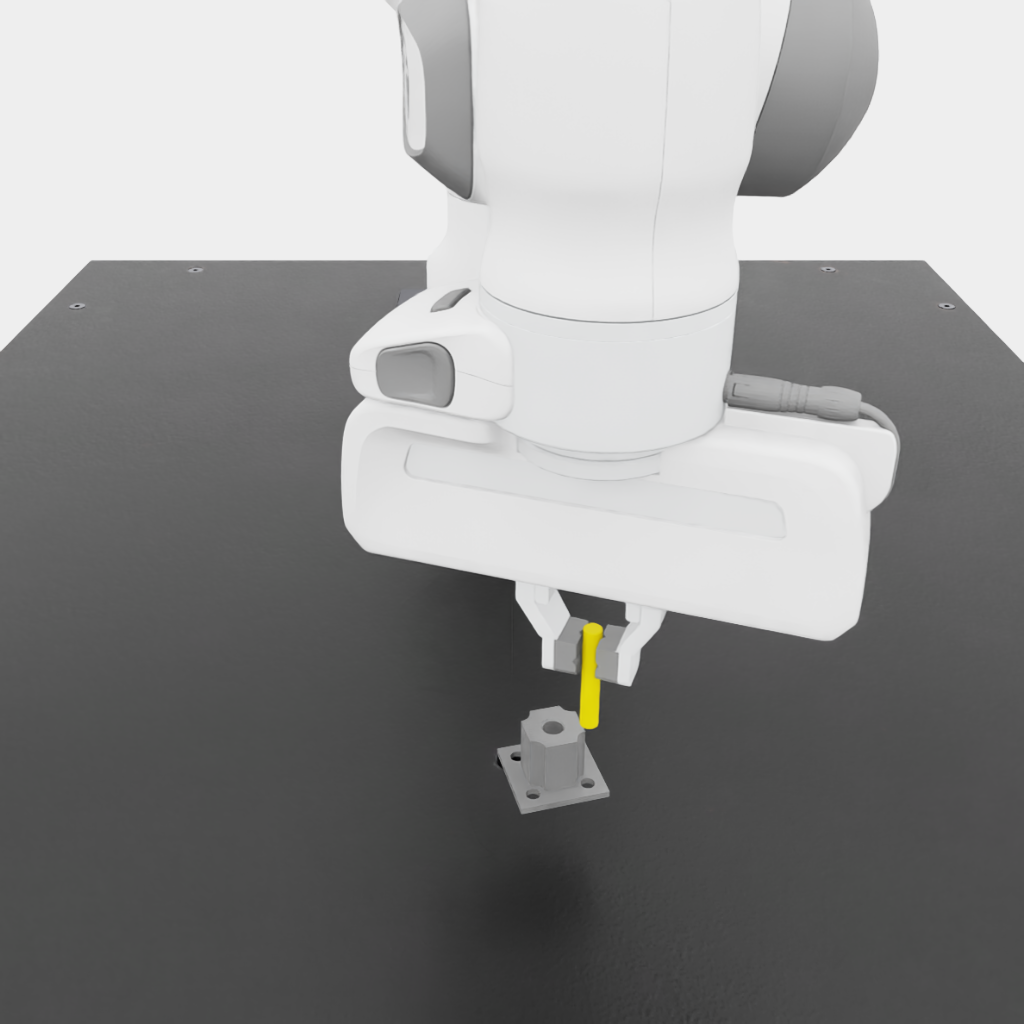} & \sceneimg{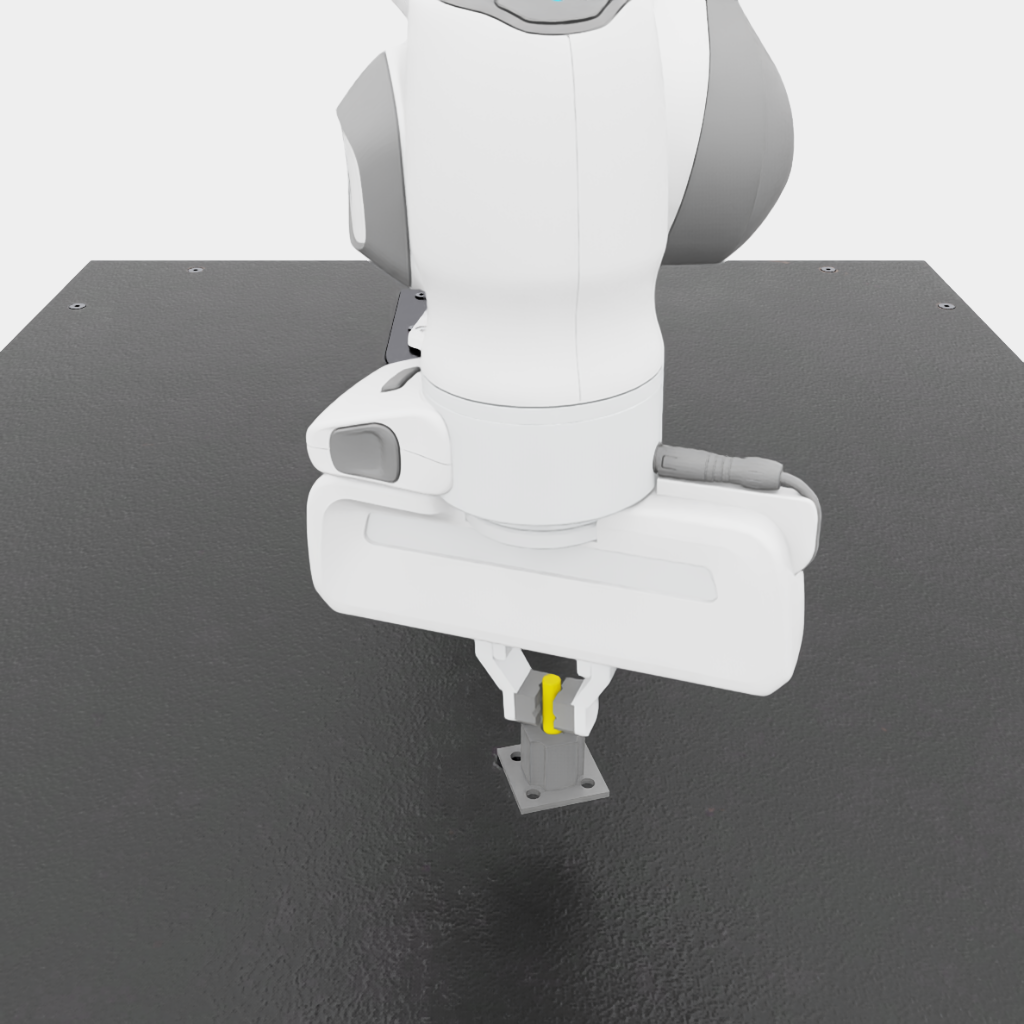} & \sceneimg{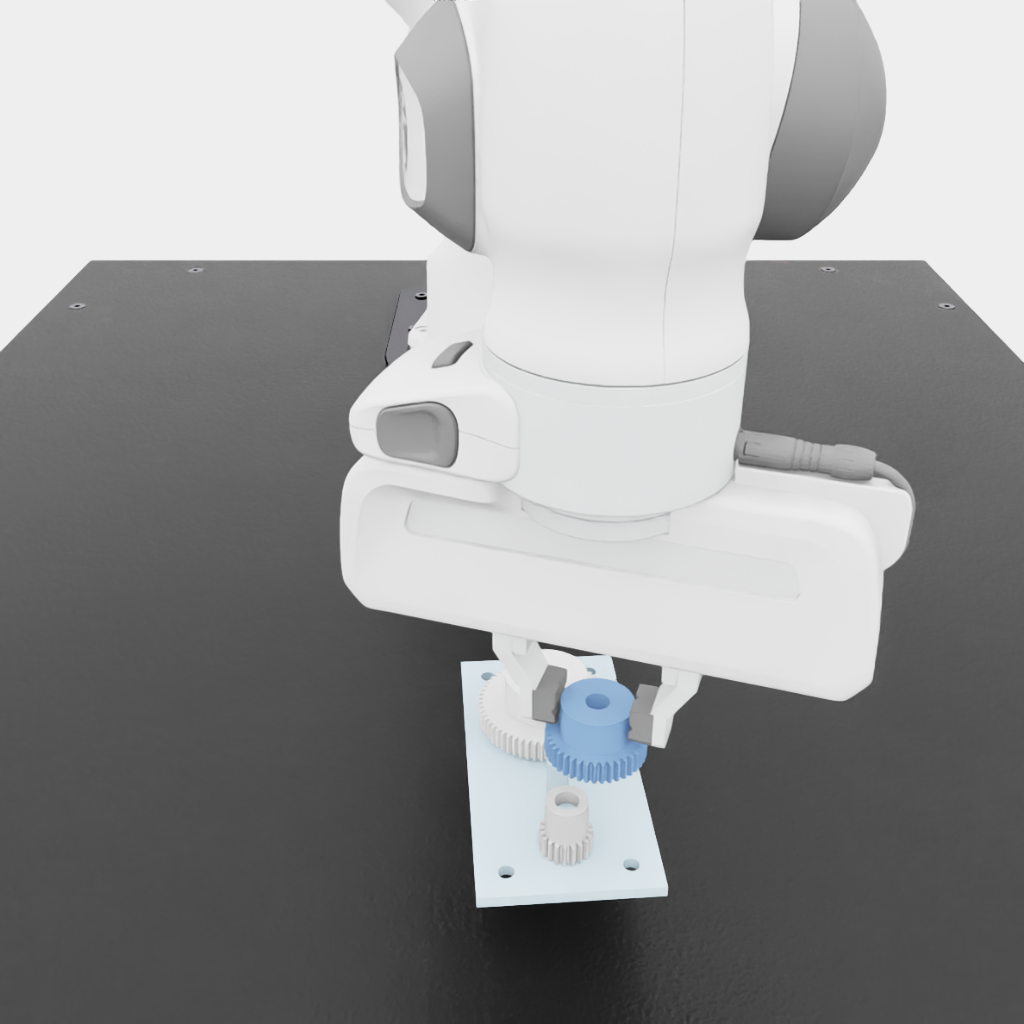} & \sceneimg{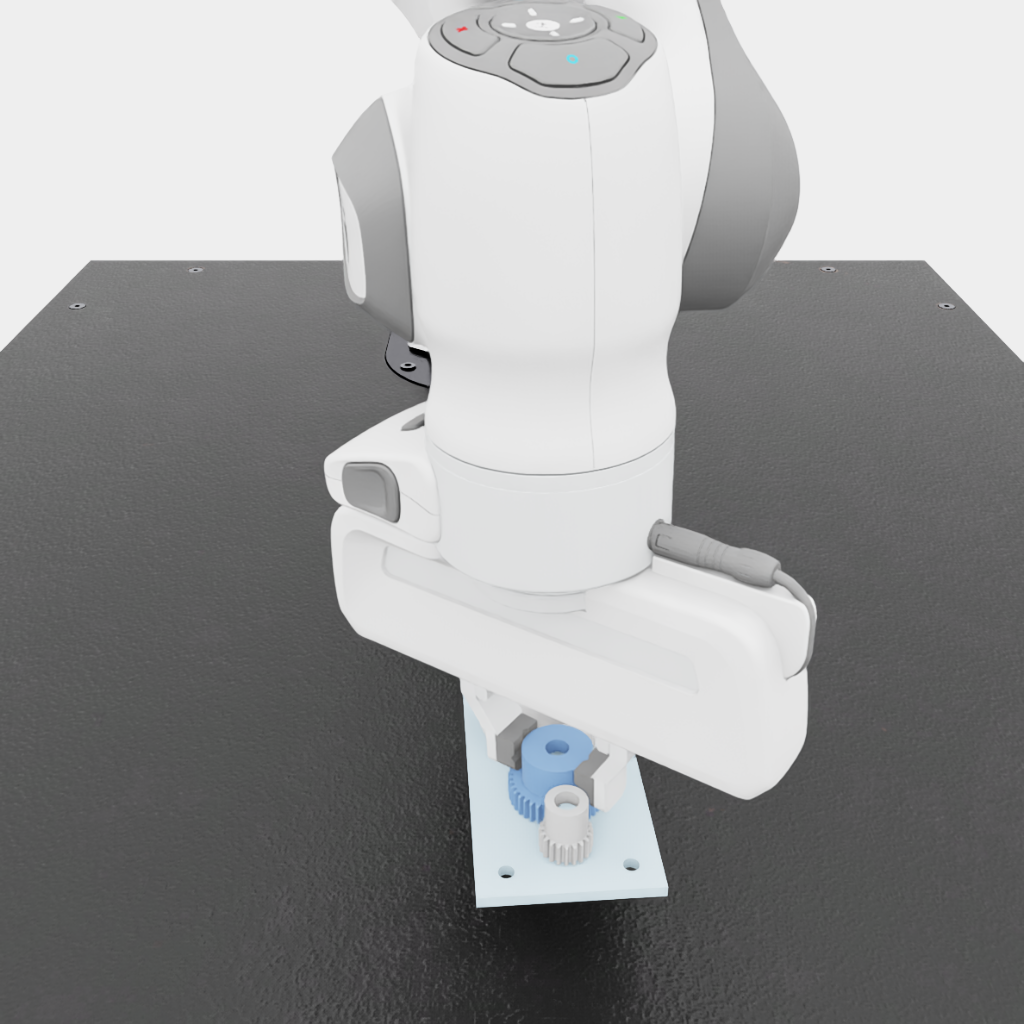} & \sceneimg{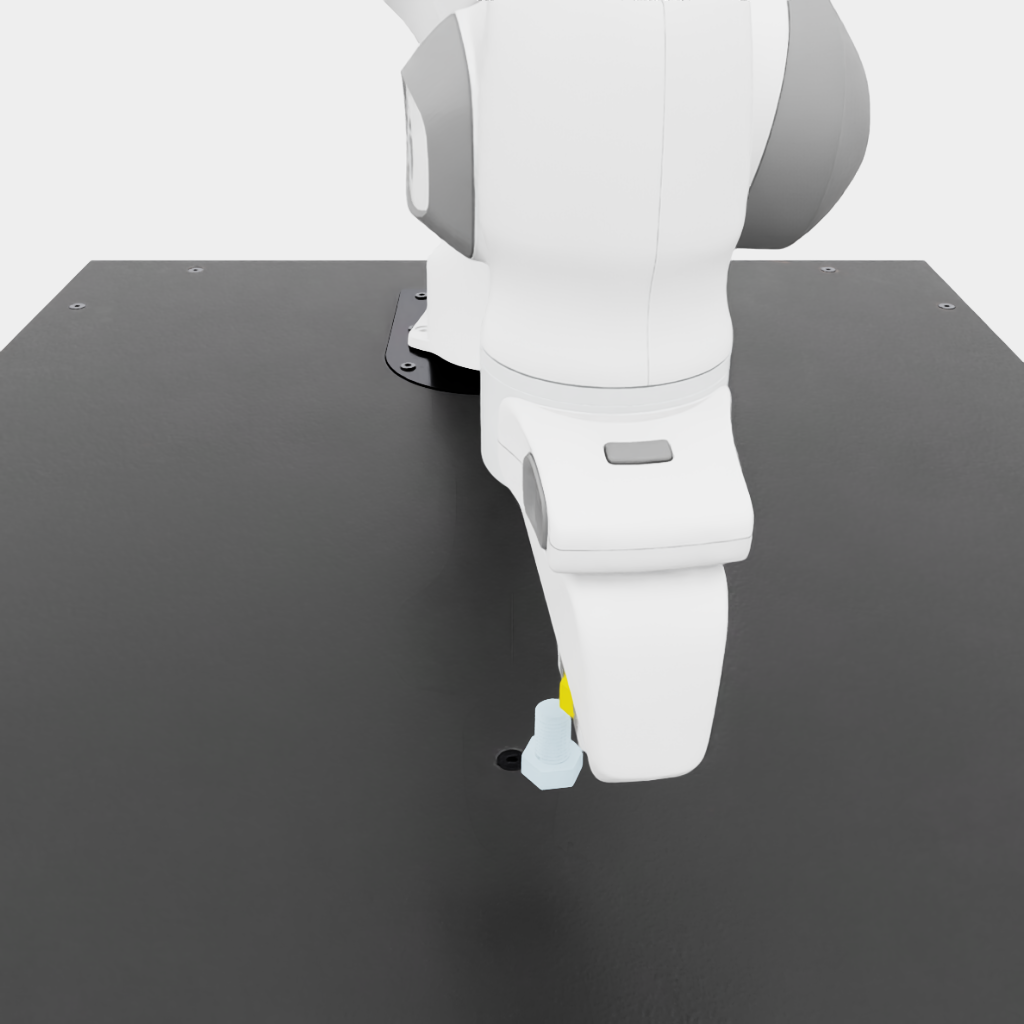} & \sceneimg{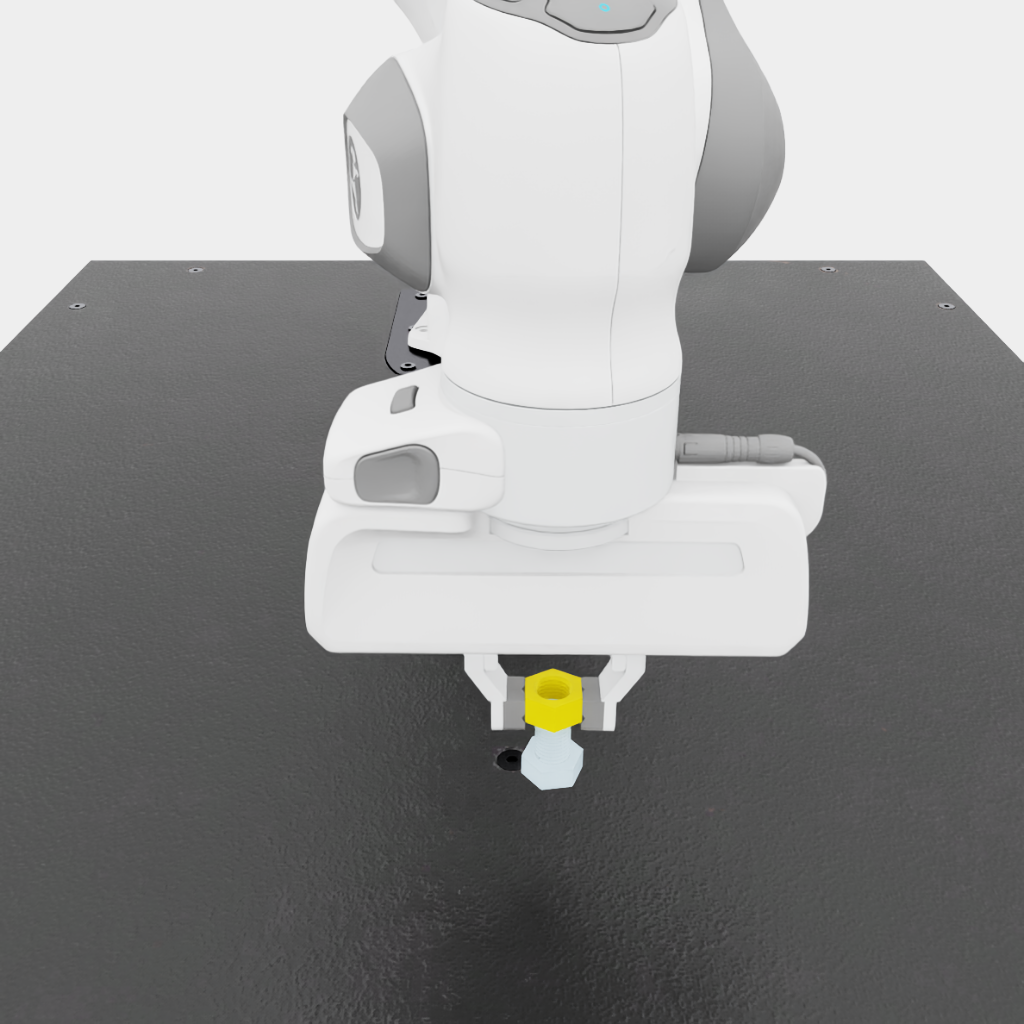} \\[1pt]
\rotatebox{90}{\makebox[0.148\textwidth]{Franka}} & \sceneimg{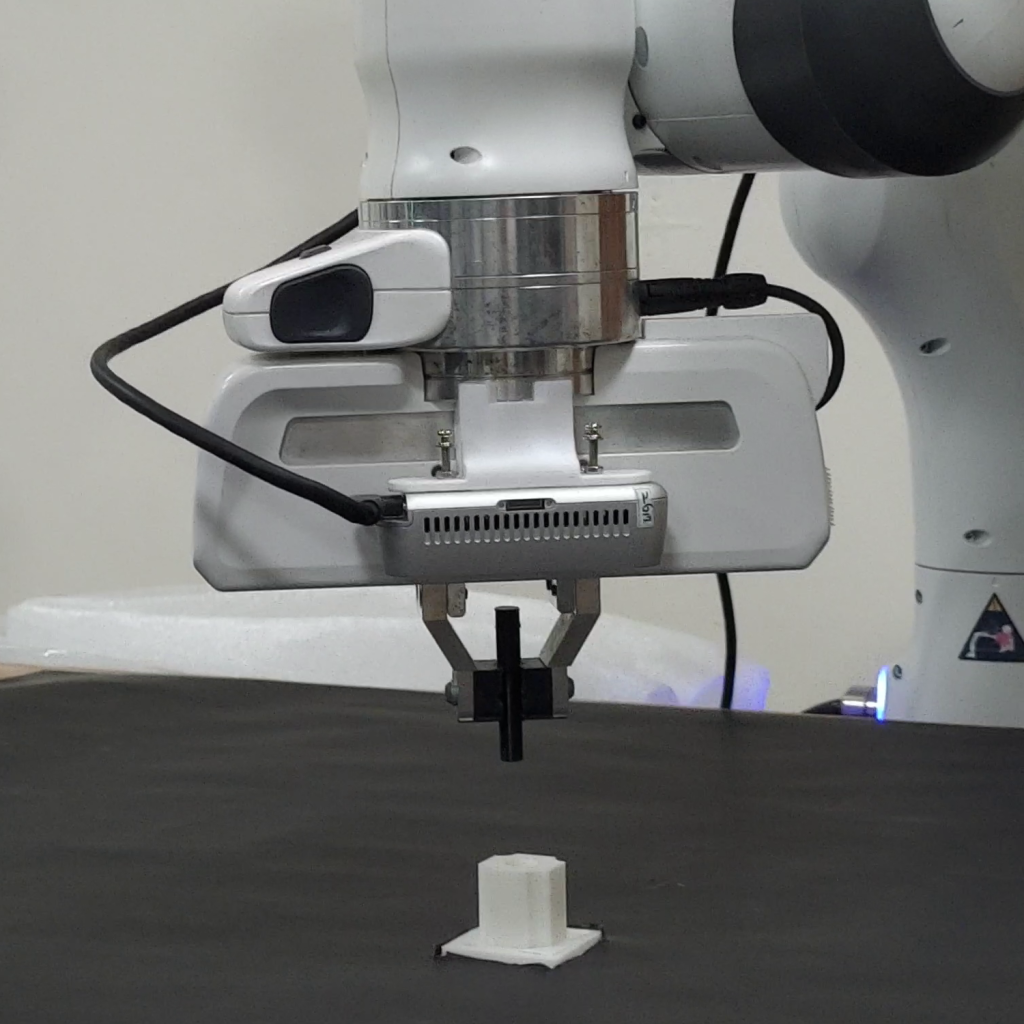} & \sceneimg{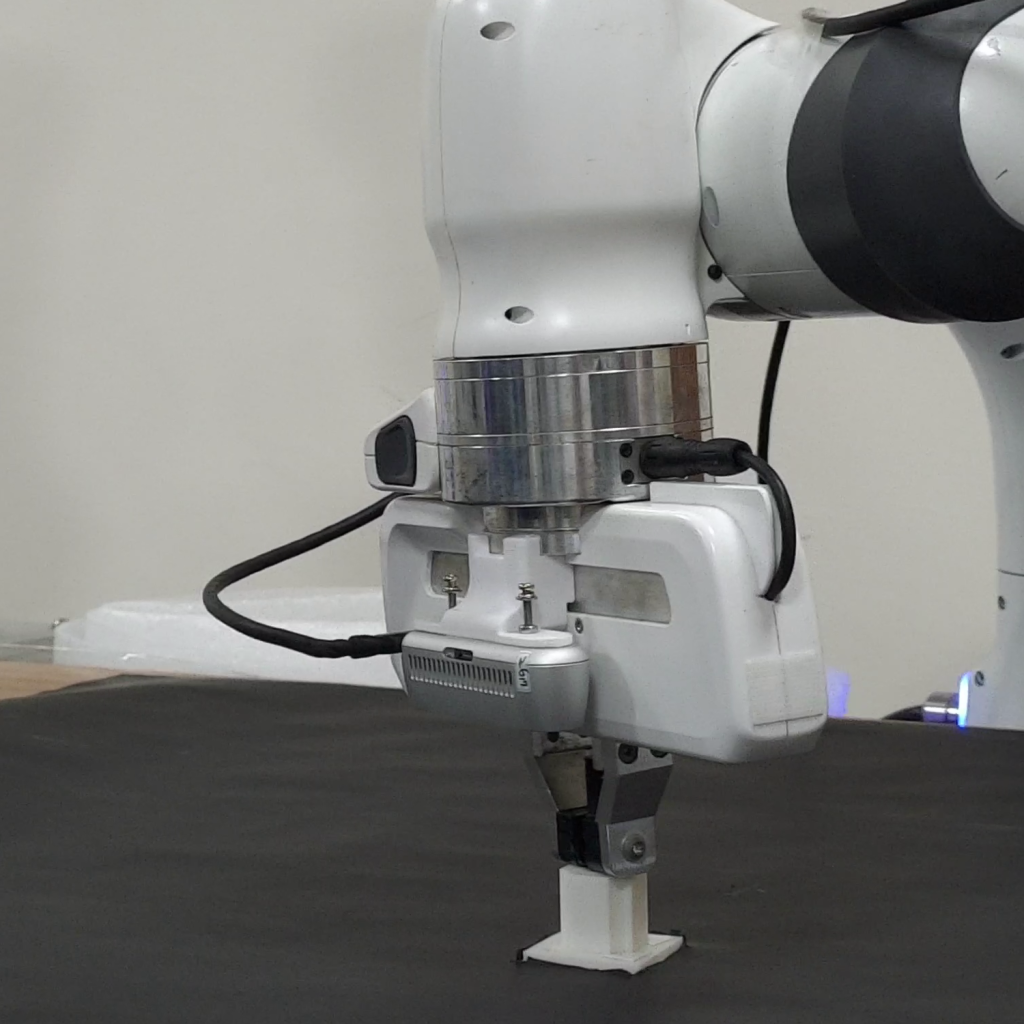} & \sceneimg{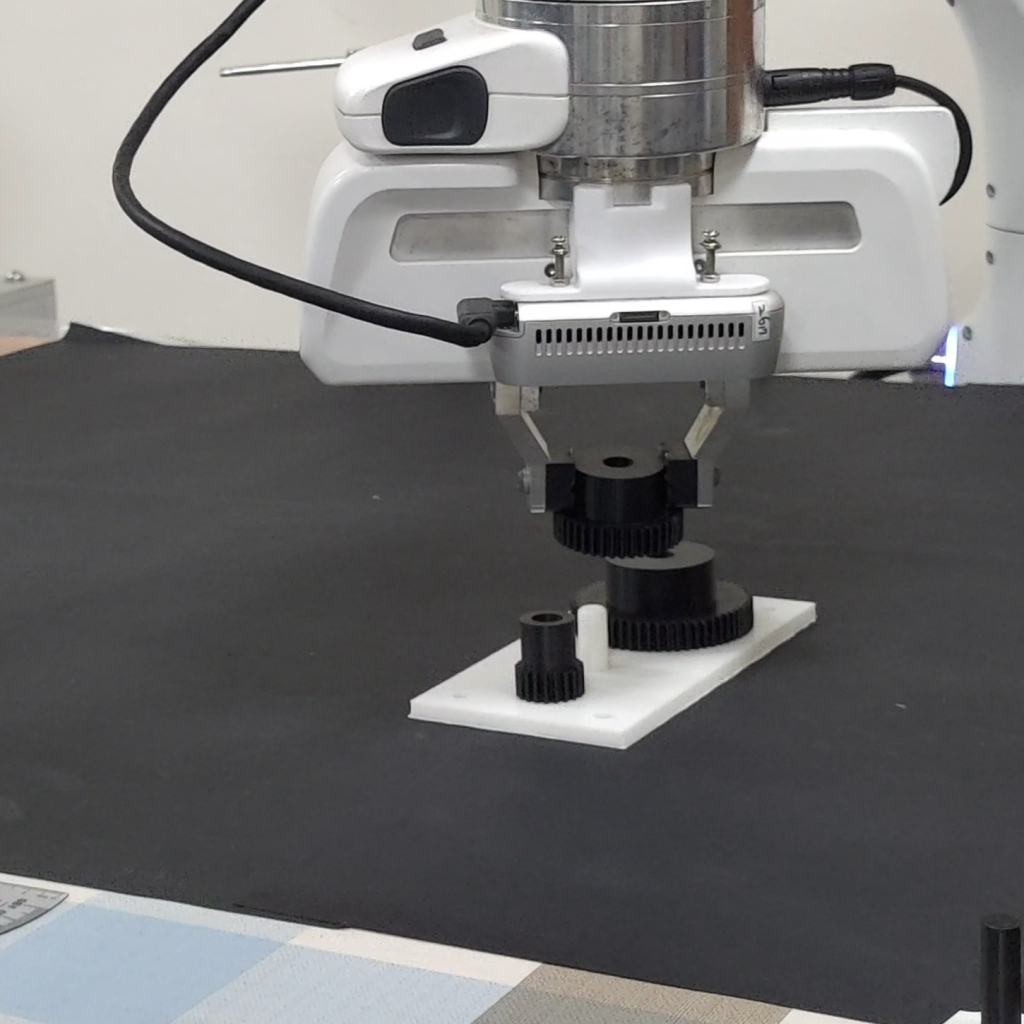} & \sceneimg{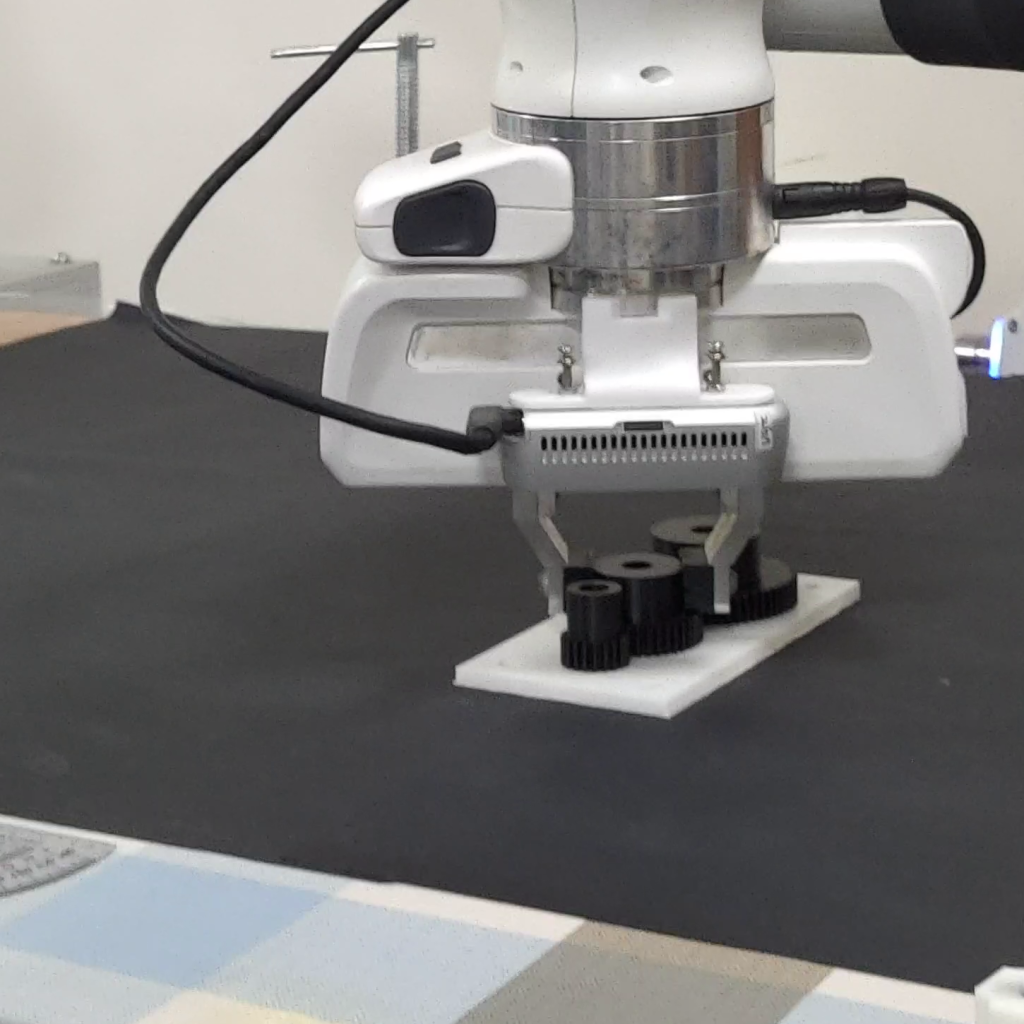} & \sceneimg{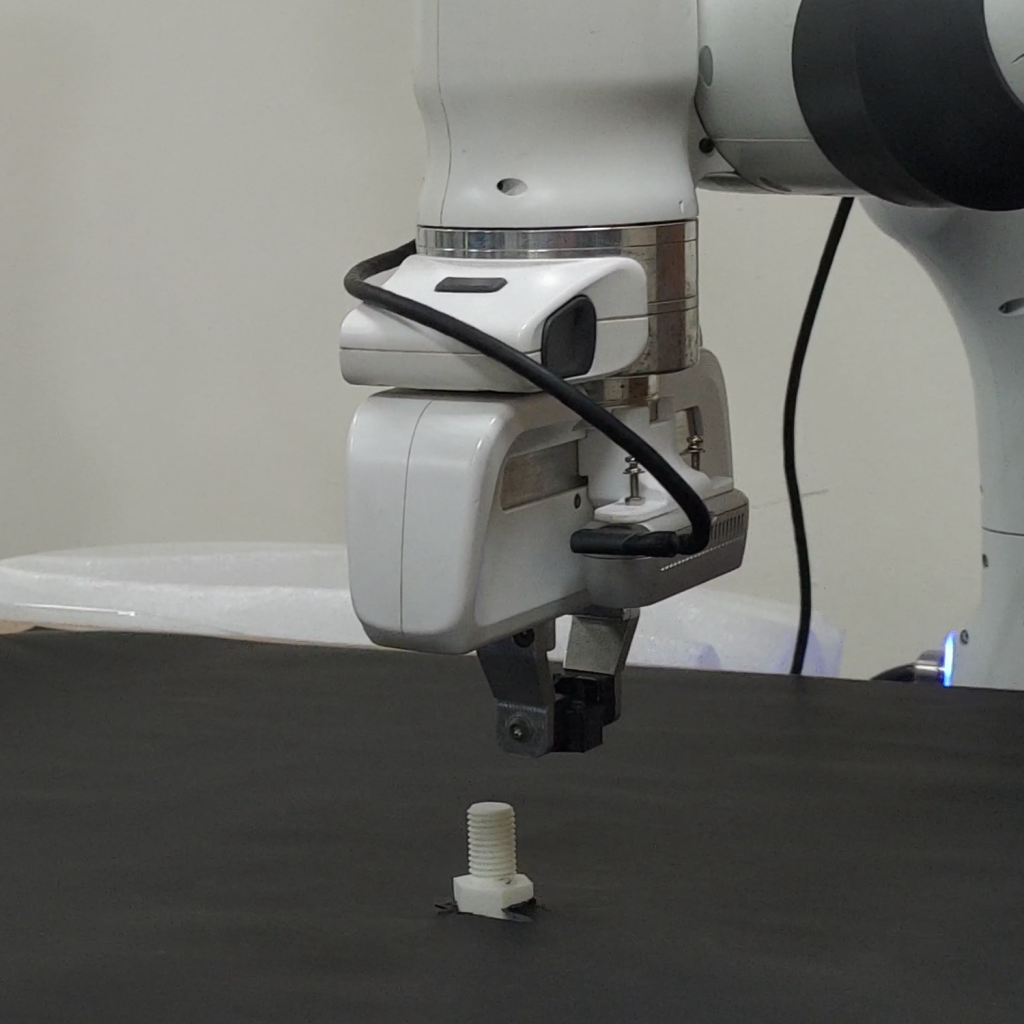} & \sceneimg{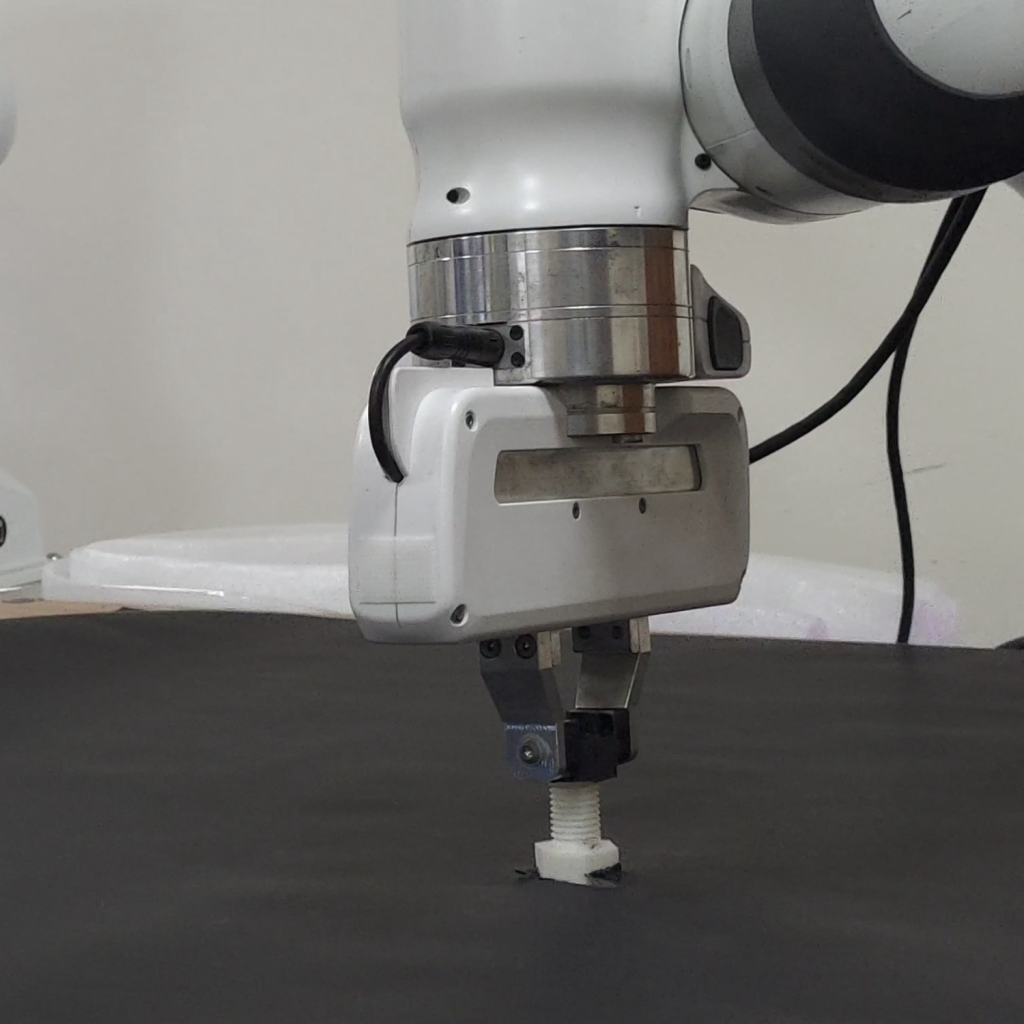} \\
\end{tabular}}
\caption{The three tasks at their initial and goal states, in simulation (top, fixed-camera view) and on the Franka (bottom). Part colours and lighting differ between simulation and the printed setup, with photometric augmentation reducing the appearance gap. Pose offsets are injected into the taught fixture pose, while the physical parts remain fixed across trials.}
\label{fig:setup}
\end{figure*}

\subsection{The robustness gain is not specific to the fusion architecture on peg}
\label{sec:arch}

On peg, three fusion variants were trained with three seeds each: bidirectional cross-attention (Fig.~\ref{fig:arch}), mean-pooled concatenation, and Action Chunking with Transformers (ACT)-style joint self-attention. At $\sigma=5$\,mm they reach $96.6{\pm}0.9\%$, $97.5{\pm}1.1\%$, and $96.7{\pm}1.3\%$, respectively, and $98.2\%$ to $98.7\%$ at $\sigma=0$. Their differences are within one pooled standard deviation, with comparable latency ($p95=3.4$\,ms on an RTX~3090). The robustness gain on peg is therefore not specific to the fusion architecture.

\subsection{Extrapolation beyond the training range}

Beyond the trained range (Fig.~\ref{fig:curves}), peg success falls to $94.5/88.8/74.2/55.5\%$ at $\sigma=7.5/10/15/20$\,mm (single seed beyond $10$\,mm), with similar trends on gear and nut. The oracle also degrades. The student-to-oracle success ratio is $97\%$ to $101\%$ within the training range, $86\%$ to $96\%$ at $2\times$, and $56\%$ to $74\%$ at $4\times$, while T-A reaches single digits and T-A+ only $3\%$ to $11\%$.

\subsection{Effect of DAgger}
\label{sec:dagger}

The paired three-seed peg comparison yields gains of $+2.7$ points at $\sigma=0$, $+7.9$ at $5$\,mm, and $+18.1$ at $10$\,mm. The increasing benefit with pose noise is consistent with greater covariate shift. At $\sigma=7.5$\,mm, cross-seed standard deviation also falls from $10.8$ to $3.2$ points.

\section{Toward the Real Robot}
\label{sec:real}

\textbf{Setup.} We deploy on a Franka Panda with Cartesian impedance control at $15$\,Hz, printed parts on a dark work surface, and the same camera crops used in training. Transfer is zero-shot: the simulation checkpoint runs unchanged, with no real-world data or fine-tuning. The student, trained with the photometric augmentation described below, and T-A+ (seed~0, our strongest camera-free policy) use identical hardware, task setup, and gating. We run $30$ trials per task and noise level at $\sigma=0$, $1$, $2.5$, and $5$\,mm. Each trial draws a fresh fixture-pose perturbation at the stated $\sigma$ and applies it to both the observation and action frame exactly as in simulation, and initial poses are randomised by $\pm20$\,mm in $xy$, $\pm10$\,mm in $z$, and $\pm45^\circ$ yaw. Figure~\ref{fig:setup} shows the initial and goal states of all three tasks in simulation and on hardware, and Table~\ref{tab:real} reports the results.

\textbf{Hardware results.} The student wins or ties every cell. Pooled across tasks ($n{=}90$ per noise level), success is $92.2\%$, $91.1\%$, $85.6\%$, and $83.3\%$ at $0$, $1$, $2.5$, and $5$\,mm, versus $88.9\%$, $85.6\%$, $68.9\%$, and $34.4\%$ for T-A+. The margin grows with pose error: $17$ points at $2.5$\,mm (Fisher's exact test, $p{=}0.012$) and $49$ points at $5$\,mm (Wilson $95\%$ intervals $[74,90]$ versus $[25,45]$, $p{<}10^{-10}$), with the largest task-level gap on peg ($80.0\%$ versus $20.0\%$).

\textbf{Factors affecting sim-to-real transfer.} Photometric augmentation \cite{tobin2017domain} is important for transfer: at $5$\,mm, the unaugmented student reaches $6\%$ to $50\%$ across tasks, compared with $84\%$ to $97\%$ after augmentation. Scene composition remains a limitation: on peg, replacing the dark table with a light one reduces success from $93\%$ to $25\%$. Camera extrinsic error also matters: increasing it from $5$\,mm$/1^\circ$ to $10$\,mm$/2^\circ$ roughly halves gear success. Two hardware interface errors, $F_z$ zero drift and a mid-episode quaternion sign flip, affect the student by at most $5$ points and do not affect the baseline.

\begin{table}[t]
\caption{Real-robot success rate (\%, $30$ trials per cell) on a Franka Panda with printed parts and a dark work surface. Offsets of the stated $\sigma$ enter both the observation and action frame, matching simulation. T-A+ is the seed-0 noise-augmented state baseline without cameras. Pooled rows combine all three tasks ($n=90$).}
\label{tab:real}
\centering
\scriptsize
\setlength{\tabcolsep}{4pt}
\begin{tabular*}{\columnwidth}{@{\extracolsep{\fill}}llcccc@{}}
\toprule
task & policy & $0$\,mm & $1$\,mm & $2.5$\,mm & $5$\,mm \\
\midrule
\multirow{2}{*}{peg}  & T-A+ & 86.7 & 83.3 & 63.3 & 20.0 \\
                      & \textbf{student} & \textbf{90.0} & \textbf{93.3} & \textbf{86.7} & \textbf{80.0} \\
\midrule
\multirow{2}{*}{gear} & T-A+ & 90.0 & 90.0 & 73.3 & 36.7 \\
                      & \textbf{student} & \textbf{93.3} & \textbf{90.0} & \textbf{83.3} & \textbf{86.7} \\
\midrule
\multirow{2}{*}{nut}  & T-A+ & 90.0 & 83.3 & 70.0 & 46.7 \\
                      & \textbf{student} & \textbf{93.3} & \textbf{90.0} & \textbf{86.7} & \textbf{83.3} \\
\midrule
\multirow{2}{*}{pooled} & T-A+ & 88.9 & 85.6 & 68.9 & 34.4 \\
                      & \textbf{student} & \textbf{92.2} & \textbf{91.1} & \textbf{85.6} & \textbf{83.3} \\
\bottomrule
\end{tabular*}
\end{table}

\section{Limitations}
\label{sec:limits}

\textbf{Scope.} The main results are simulated, and the hardware sweep has $30$ trials per cell. Table~\ref{tab:main}, the architecture ablation, the DAgger comparison, and the appearance experiments use three seeds. In Table~\ref{tab:modality}, three-seed cells are marked explicitly and the remainder are single-seed. The external baselines, non-privileged gear and nut arms, and relabelled $+$DAgger arm are also single-seed. Beyond $5$\,mm, T-A, T-A+, and T-B are single-seed, while the peg student becomes single-seed only beyond $10$\,mm. Extending the matched-law baseline from T-A's $200$-epoch budget to $1.5\times$ raises peg success at $5$\,mm only from $31.6\%$ to $39.5\%$.

\textbf{Geometry.} The student is trained only on the three FORGE geometries. On $70$ AutoMate \cite{tang2024automate} plug-socket pairs in the unmodified peg environment ($\sigma=2.5$\,mm, $n=32$ per cell), it succeeds in $4.6\%$ of trials versus $18.4\%$ for the noise-augmented state baseline. Failures are dominated by visual mis-localisation of unseen geometries, which directs the student toward incorrect target locations. Robustness to pose error therefore does not imply cross-geometry generalisation. The $\epsilon$-regressor shows the same limitation and has not been evaluated on gear, nut, or hardware.

\section{Conclusion}

Pose-estimation error in precision assembly is not merely sensing noise: when the same estimate also defines the action frame, it creates a latent offset that proprioception cannot resolve before contact. Encoding compensation through a privileged teacher or closed-form relabelling yields a deployable student that maintains $92\%$ to $99\%$ success through $\sigma=5$\,mm and recovers $93\%$ to $100\%$ of the oracle's advantage. Zero-shot on a Franka, it reaches $83.3\%$ versus $34.4\%$ for the strongest state-based policy. The same architecture reaches only $31.5\%$ when trained from demonstrations without offset access, versus $88.7\%$ and $95.8\%$ with privileged and relabelled demonstrations. Robustness therefore comes from supervision that teaches compensation for the displaced action frame.

\section*{Acknowledgment}
An AI language model (Claude, Anthropic) was used to polish the wording of this
paper under the authors' direction. All experiments, numbers and claims were
produced and verified by the authors.

\end{document}